\documentclass{article}

\usepackage[colorlinks,citecolor=blue,bookmarks=false]{hyperref}
\usepackage{geometry}

\usepackage[utf8]{inputenc}
\usepackage{amsmath,amssymb}
\usepackage{graphicx} 
\usepackage{subcaption}
\usepackage{xspace}
\usepackage{color}
\usepackage{soul}
\usepackage{float}
\usepackage{booktabs}\let\cline\cmidrule
\usepackage{multirow}
\usepackage{algorithm}
\usepackage[noend]{algpseudocode}
\usepackage{listings}
\usepackage{natbib}
\usepackage{enumitem}
\usepackage{tikz}
\usepackage{enumitem}
\usetikzlibrary{bayesnet}
\usepackage{arydshln}
\usepackage{mathtools}

\newcommand{\nE}{\mathbb{E}}

\newcommand{\nI}{\mathbb{I}}

\newcommand{\nN}{\mathbb{N}}

\newcommand{\nR}{\mathbb{R}}

\newcommand{\cD}{\mathcal{D}}

\newcommand{\cM}{\mathcal{M}}
\newcommand{\cN}{\mathcal{N}}

\newcommand{\cU}{\mathcal{U}}

\newcommand{\figref}[1]{Fig.~\ref{#1}}
\newcommand{\secref}[1]{Section~\ref{#1}}
\newcommand{\eqnref}[1]{Eq.~\eqref{#1}}
\newcommand{\tabref}[1]{Table~\ref{#1}}
\newcommand{\algoref}[1]{Algorithm~\ref{#1}}

\makeatletter
\DeclareRobustCommand\onedot{\futurelet\@let@token\@onedot}
\def\@onedot{\ifx\@let@token.\else.\null\fi\xspace}

\makeatother

\def\ws{\omega_T}
\def\ys{y_T}

\def\Ns{N_T}

\def\we{\omega_I}
\def\ye{y_I}
\def\se{\sigma_I}

\def\Ne{N_I}

\newcommand*{\getTitle}{Uncertainty Quantification and Propagation in Surrogate-based Bayesian Inference}
\newcommand*{\getFirstAddress}{Cluster of Excellence SimTech, University of Stuttgart, Germany}
\newcommand*{\getSecondAddress}{Department of Statistics, TU Dortmund University, Germany}
\newcommand*{\getFirstKeyword}{Surrogate Modeling}
\newcommand*{\getSecondKeyword}{Uncertainty Quantification}
\newcommand*{\getThirdKeyword}{Uncertainty Propagation}
\newcommand*{\getFourthKeyword}{Bayesian Inference}
\newcommand*{\getFifthKeyword}{Machine Learning}

\title{\getTitle}
\author{Philipp Reiser$^{1,*}$, Javier Enrique Aguilar$^{1,2}$, Anneli Guthke$^{1}$, Paul-Christian Bürkner$^{1,2}$}
\date{
    $^1$\getFirstAddress\\
    $^2$\getSecondAddress\\
    $^*$Corresponding author, Email: philipp-luca.reiser@simtech.uni-stuttgart.de
}

\begin{document}

\maketitle
\begin{abstract}
    Surrogate models are statistical or conceptual approximations for more complex simulation models. In this context, it is crucial to propagate the uncertainty induced by limited simulation budget and surrogate approximation error to predictions, inference, and subsequent decision-relevant quantities. However, quantifying and then propagating the uncertainty of surrogates is usually limited to special analytic cases or is otherwise computationally very expensive. In this paper, we propose a framework enabling a scalable, Bayesian approach to surrogate modeling with thorough uncertainty quantification, propagation, and validation. Specifically, we present three methods for Bayesian inference with surrogate models given measurement data. This is a task where the propagation of surrogate uncertainty is especially relevant, because failing to account for it may lead to biased and/or overconfident estimates of the parameters of interest. We showcase our approach in three detailed case studies for linear and nonlinear real-world modeling scenarios. Uncertainty propagation in surrogate models enables more reliable and safe approximation of expensive simulators and will therefore be useful in various fields of applications.

\end{abstract}

\section{Introduction}\label{section:introduction}
\begin{figure*}
    \begin{center}
    \includegraphics[width=\textwidth]{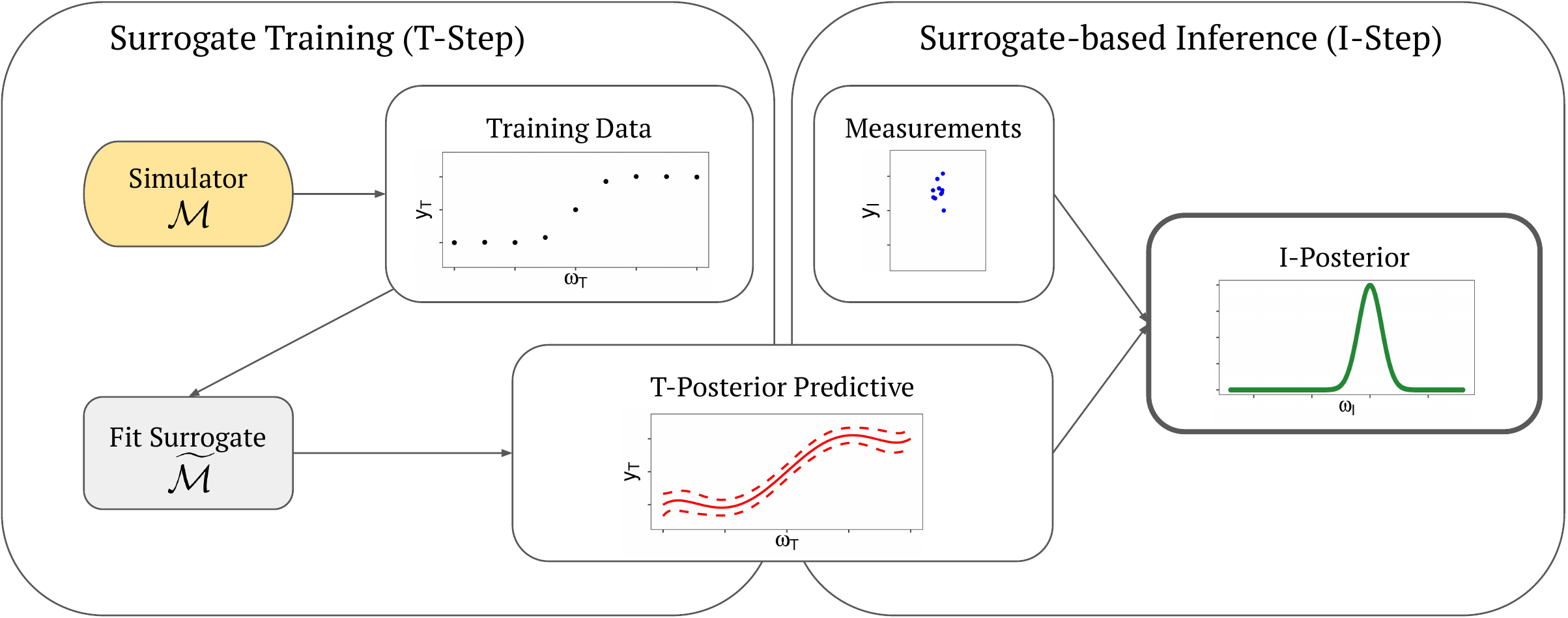}
    \caption{Overview of two-step procedure. Left: In the surrogate training step (T-Step), training data is generated using a simulator and a surrogate model is fitted which allows to estimate the T-posterior. Right: In the surrogate-based inference step (I-Step), measurements along with the T-posterior are used to infer the I-posterior.}
    \label{fig:overview}
    \end{center}
\end{figure*}
Simulations of complex phenomena are crucial in the natural sciences and engineering for different scenarios, e.g., for gaining system understanding, prediction of future scenarios, risk assessment, or system design. However, often they are based on complex ordinary differential equations or partial differential equations which may not have closed-form solutions and may have to be solved using expensive numerical methods. To overcome computational overhead, the field of surrogate models \citep{ZHU2018surrconvendec, gramacy2020surrogates, lavin2021simint} has emerged which provide fast approximations of computationally expensive simulation. Examples are polynomial chaos expansion \citep{Wiener1938JHUP, Sudret2008RESS, Oladyshkin2019RESS, Buerkner2022bspce}, Gaussian processes \citep{kennedy2001bayesiancalibration, rasmussen2005gpforml} or neural networks \citep{Goodfellow-et-al-2016}. Recently, there has been a great interest in applying surrogate models in relevant areas, for example in hydrology \citep{MOHAMMADI201853, tarakanov_regression-based_2019, Zhang2020WRR}, in fluid dynamics \citep{meyer2021fluid}, in climate prediction \citep{kuehnert_surrogate_2022}, or in systems biology \hbox{\citep{renardy2018uqsurrsysbio, alden2020usingemulation}}. Furthermore, great methodological advances have been made in the field of surrogate modeling, for example, the incorporation of physical knowledge \citep{raissi2019physicsinformednn, li2021fno, brandstetter2023clifford} or the combination of surrogate models and simulation-based inference \citep{radev2023jana}.

Despite these advances, a major remaining challenge is the trustworthiness and reliability of the surrogate. Consequently, it is crucial to quantify uncertainties associated with surrogate modeling, e.g., caused by limited training data or inflexibility of the surrogate. To estimate the uncertainty in surrogate model parameters, several methods for uncertainty quantification (UQ) have been developed \citep[e.g.,][]{bsparsepceforgs2017shao, Buerkner2022bspce}. Uncertainty propagation (UP), a sub-field of UQ, is particularly important for addressing surrogate uncertainties in subsequent (surrogate-based) inference tasks \citep{smith2013uq, lavin2021simint, psaros2023uqinsciml}.

Using probability theory as its fundamental basis, Bayesian statistics provides a rigorous way for UQ generally and specifically for UP \citep{gelman2013bayesian, mcelreath2020statistical, buerkner2022models}. 
One important use-case for UP in surrogate modeling arises from the ``forward'' problem, in which input parameters are uncertain and the goal is to propagate them through the surrogate while accounting for its uncertainty to compute a reliable output. For example, \cite{ranftl2021bayesian} proposed a method for  UP in the forward problem, for restricted cases where conjugate analytic models are available. Similarly, \cite{ZHU2018surrconvendec} quantified uncertainty in neural network surrogate outputs using approximate Bayesian inference. 

In the ``inverse'' problem, when surrogates are used for Bayesian inference of parameters given observed data \citep[e.g.,][]{kennedy2001bayesiancalibration, marzouk2007stochasticsprectral, marzouk2009stochasticcollocation, zeng2012sparsegridbayesian, laloy2013effposteriorexpl, li2014adaptiveconstruction, clearly2021calibrateemulatesample}, rigorous propagation of the surrogate uncertainty becomes even more relevant and challenging.

The framework introduced by \cite{kennedy2001bayesiancalibration}, for instance, attempts to jointly infer unknown input and surrogate parameters using data from both simulation models and observations. However, this approach leads to a loss of control over which parameters are updated by which data source, also implying complex posteriors that are hard to sample from \citep{bayarri2009modularizationbayesian}.

To address these issues, modularization \citep{bayarri2009modularizationbayesian} has been introduced to surrogate modeling, aiming to update only specific parameters using selected data. 
Despite these advances, the uncertainty of the surrogate parameters is often neglected or simplified in the context of surrogate modeling. For example, \cite{bayarri2009modularizationbayesian} propagated only point estimates of the surrogate parameters between modules, thus neglecting relevant uncertainty in subsequent calculations.
Further, \cite{Zhang2020WRR} applied surrogate-based Bayesian inference to hydrological systems and propagated parts of the surrogate uncertainty assuming normal surrogate posteriors. We review these methods in more detail in \secref{subsubsection:E-Lik}.

In this paper, our primary focus lies on UP when solving probabilistic inverse problems via surrogate models. 
Existing methods proposed for the same challenge are scarce and propagate the surrogate-based uncertainty only in a selective and simplified manner, which leaves a lot of room for both improved theory and improved practical methods.
This not only concerns UP itself, but also diagnostic methods to assess the validity of the resulting inference. This paper aims at addressing these challenges from a fully probabilistic (Bayesian) perspective.
Concretely, we make the following contributions: (i) Within a formal framework for surrogate-based Bayesian inference, we specify and categorize all relevant uncertainties.
(ii) We present three distinct methods for propagating surrogate uncertainty within a two-step inference procedure, which consists of a surrogate training step (T-Step) and a surrogate-based inference step (I-Step). For a high-level overview, see \figref{fig:overview}. (iii) We adapt existing simulation-based procedures to validate the uncertainty calibration achieved via the different UP approaches. Finally, we evaluate our methods in three detailed case studies.

\section{Method}\label{section:method}

In the following, we propose a framework for uncertainty propagation (UP) in surrogate-based Bayesian inference. This framework is applicable to general surrogate models, e.g., linear regression, polynomial chaos expansion, Gaussian processes or neural networks. The framework consists of a two-step procedure with a surrogate training step (T-Step) and an inference step (I-Step), an overview of all uncertainties occurring in surrogate-based inference, several UP schemes of selected uncertainties, and their evaluation.

\subsection{Two-Step Procedure}\label{subsection:2-step-procedure}
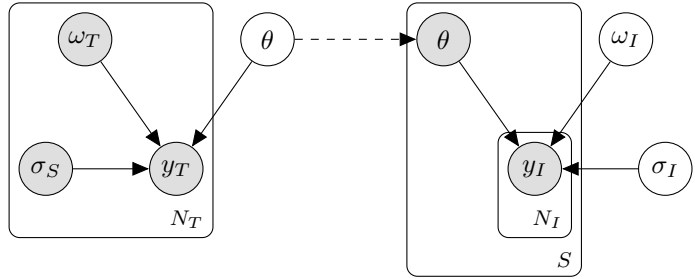
\begin{figure*}[ht]
  \begin{center}
    \begin{tabular}{cc}
      \begin{tikzpicture}

  \node[obs]                               (y_T) {$y_T$};
  \node[obs, above=of y_T, xshift=-1.2cm]    (w_T) {$\omega_T$};
  \node[obs, left=of y_T]    (sigma_S) {$\sigma_S$};
  \node[latent, above=of y_T, xshift=1.2cm]  (theta) {$\theta$};
  \node[obs, right=4cm of y_T]               (y_I) {$y_I$};
  \node[latent, above=of y_I, xshift=1.2cm]    (w_I) {$\omega_I$};
  \node[obs, above=of y_I, xshift=-1.2cm]  (theta_S) {$\theta$};
  \node[latent, right=of y_I]    (sigma_I) {$\sigma_I$};

  \edge {w_T, theta, sigma_S} {y_T} ; 
  \edge {w_I, theta_S, sigma_I} {y_I} ; 
  \edge[dashed] {theta} {theta_S} ;

  \plate {y_Tw_T} {(w_T)(y_T)(sigma_S)} {$N_T$} ;
  \plate {y_I} {(y_I)} {$N_I$} ;
  \plate {y_Itheta_S} {(theta_S)(y_I)} {$S$} ;
\end{tikzpicture}
    \end{tabular}
  \end{center}
  \caption{Graphical model of the T-Step and I-Step. Left: In the T-Step, the observed quantities are simulation parameters $\ws$, simulation output $\ys$, and the noise hyperparameters $\sigma_S$. The unknowns are the surrogate parameters $\theta$. Right: In the I-Step, measurement data $\ye$ is observed $N_I$ times and $S$ posterior samples of $\theta$ are propagated from the T-Step. The dashed arrow indicates that uncertainty in $\theta$ is propagated to the I-Step while $\theta$ is not updated using the data $y_I$. The unknowns to be inferred  are the simulation parameters $\we$ and the measurement error hyperparameters $\sigma_I$.}
  \label{fig:2_step_graphical_model}
\end{figure*}

We propose a two-step procedure for uncertainty propagation of surrogate models which consists of (i) the Surrogate-Training Step (T-Step), where training data is generated using a simulator and (ii) the Inference-Step (I-Step) where the trained surrogate is used to infer a quantity of interest, as illustrated in the graphical model \citep{jordan1999leaningingraphicalmodels, bengal2008bayesiannetworks} shown in \figref{fig:2_step_graphical_model}. Even though the general setup may look relatively simple, it becomes challenging to quantify and propagate all occurring uncertainties in a statistically rigorous manner. In the following, we differentiate between aleatoric (irreducible) and epistemic (reducible) uncertainty; for a review see \cite{hullermeier2021aleatoric, gruber2023sources}. In \tabref{table:uncertainties2step} we summarize all relevant uncertainties occurring in the two-step procedure to be discussed in detail below.

\begin{table*}[ht]
  \caption{Uncertainties in the two-step procedure. For each parameter and posterior distribution, we list the type of uncertainty (epistemic or aleatoric) in the T-/I-Step. We also list the synonyms used throughout the text.}
  \centering
  \renewcommand{\arraystretch}{1.2}
  \begin{tabular}{llll}
    \toprule
    \textbf{Parameter/Posterior Distribution}            & \textbf{T-Step} & \textbf{I-Step} & \textbf{Synonyms}   \\
    \midrule
    Simulator noise                 & aleatoric &   -          & - \\
    hyperparameter $\sigma_S$       &           &              & \\ \hdashline
    Surrogate approx.               & aleatoric & aleatoric    & T-aleatoric uncertainty \\
    error hyperparameter $\sigma_A$ &           &              & \\ \hdashline
    Surrogate parameter             & epistemic & aleatoric    & T-epistemic uncertainty / \\ 
    posterior $p(c, \sigma_A \mid \cD_T)$     &           &              & T-posterior \\ \hdashline
    Measurement noise               & -         & aleatoric    & -       \\
    hyperparameter $\sigma_I$        &           &              &  \\ \hdashline
    Simulator-based                 & -         & epistemic    & -\\
    posterior $p(\omega_I \mid y_I; \cM)$&           &              & \\ \hdashline
    Surrogate-based                 & -         & epistemic \& & I-posterior\\
    posterior $p(\omega_I \mid y_I; \widetilde{\cM}, u)$ & &  aleatoric\textsuperscript{*}& \\
    \bottomrule
  \end{tabular}
  \caption*{\textsuperscript{*} Depends on uncertainty propagation method $u$.}
  \label{table:uncertainties2step}
\end{table*}

\subsubsection{First Step: Training the Surrogate (T-Step)}\label{subsubsection:first_step}
In the first step, we focus on training a surrogate model using artificial data generated from the complex simulator we seek to approximate. This involves computing a posterior over the surrogate parameters to account for the induced uncertainties.

\paragraph{Simulator}
We consider a given (potentially stochastic) simulator $\cM$, which is an arbitrarily complex model, for example describing a physical or biological process, and  which is hard to evaluate.
Given an input simulation parameter $\omega_T$, we obtain the output response from the simulator as follows:
\begin{align}
    y_T = \cM(\omega_T; e_S) \quad \text{with} \quad e_S \sim p(e_S \mid \sigma_S),
\end{align}
where $e_S$ is the noise of the simulation drawn from a simulator noise distribution $p(e_S \mid \sigma_S)$ with hyperparameters $\sigma_S$ that describe the aleatoric (irreducible) uncertainty of the simulation. Note, that the simulation model itself does not necessarily have to be stochastic; it could be a deterministic physics-based or conceptual model that is complemented with a stochastic representation of measurement noise. We explicitly choose the subscript $S$ to highlight that the noise stems from the simulation and not from the training process. This setup induces the true generating distribution $p(y_T \mid \omega_T, \sigma_S)$ of simulation responses $y_T$ given input simulation parameters $\omega_T$ and noise hyperparameters $\sigma_S$.

\paragraph{Surrogate Model}
A surrogate model $\widetilde{\cM}$ is a statistical model that aims to approximate the simulator $\cM \approx \widetilde{\cM}$ while being computationally more efficient to evaluate. The parametric form of the surrogate is defined by a set of surrogate approximation parameters $c$. For example, if the surrogate model is a polynomial, then $c$ are the polynomial coefficients. Given input parameters $\omega_T$, surrogate approximation parameters $c$, and the simulator noise distribution $p(e_S \mid \sigma_S)$, we can calculate the surrogate response:
\begin{align}
    \tilde{y}_T =\widetilde{\cM}(\omega_T, c, e_S) \quad \text{with} \quad e_S \sim p(e_S \mid \sigma_S).
\end{align}

\paragraph{Surrogate Approximation Error} 
A surrogate model with a fixed architecture will always be misspecified, if the true simulator is not included in the class of surrogate models that we define. This is the standard use case, since we are specifically tailoring the surrogate model to be much simpler than the simulator. The uncertainty of the surrogate approximation parameters $c$ only captures the (epistemic) uncertainty due to limited training data. To additionally account for the approximation error of the surrogate with respect to the simulator, caused by limited expressibility of the surrogate, we introduce an additional error term $e_A$. We assume $e_A$ to follow a distribution $p(e_A \mid \sigma_A)$ with hyperparameters $\sigma_A$. This distribution captures aleatoric uncertainty which is not reducible by more training data. We can then model the true response $y_T$ via a (potentially unknown) function $\tilde{f}$ that takes the output of the surrogate $\tilde{y}_T$ and the approximation error $e_A$:
\begin{align}
    y_T = \tilde{f}(\tilde{y}_T, e_A) \quad \text{with} \quad e_A \sim p(e_A \mid \sigma_A).
\end{align}
In practice, we might assume a simple additive error: $y_T = \tilde{y}_T + e_A$ and a normal distribution for $e_A$, but our framework is agnostic to these choices. For the sake of readability, we use $\theta=\{c, \sigma_A\}$ to combine all trainable surrogate parameters in a single vector. Together, this implies a surrogate likelihood of the simulator responses $y_T$:
\begin{equation}
    y_T \sim p(y_T \mid \omega_T, \sigma_S, \theta).
\end{equation}

\paragraph{Surrogate Training}
To train the surrogate, we use the simulator $\cM$ to generate training data $\cD_T = \{ \ws^{(i)}, \sigma_S^{(i)}, \ys^{(i)} \}_{i=1}^{\Ns}$ consisting of $N_T$ inputs $(\ws^{(i)}, \sigma_S^{(i)})$ and corresponding outputs $\ys^{(i)} \sim p(y_T^{(i)} \mid \ws^{(i)}, \sigma_S^{(i)})$. The goal of the T-Step is to fit the parameters $\theta$ of the surrogate $\widetilde{\cM}$ to approximate the data distribution implied by $\cM$.

To train our surrogate parameters $\theta$, we perform Bayesian inference using the fast-to-evaluate surrogate likelihood $p(y_T \mid \omega_T, \sigma_S, \theta)$ and a (potentially non-informative) prior $p(\theta)$. We obtain the joint posterior distribution over all surrogate parameters given the simulation training data $\cD_T$ as:
\begin{align}
    p(\theta \mid \cD_T) \propto \prod_{i=1}^{N_T}p(y_T^{(i)} \mid \omega_T^{(i)}, \sigma_S^{(i)}, \theta) \, p(\theta).
\end{align}
This joint posterior describes the epistemic (reducible) uncertainty in the surrogate model parameters. In the case of infinite training data, i.e. $N_T \to \infty$ the posterior $p(\theta \mid \cD_T)$ converges to a point mass under regularity conditions \citep{van2000asymptotic}. To approximate this posterior we can use sampling-based algorithms, such as Markov chain Monte Carlo (MCMC) \citep{robert2005monte}, and represent the posterior in the form of $S$ posterior samples $\{\theta^{(1)}, \dots, \theta^{(S)}\} \sim p(\theta \mid \cD_T)$. The framework is in principle agnostic to the choice of estimation algorithm, as long as the algorithm can be used to obtain posterior samples. This includes MCMC but also other methods such as variational inference \citep{kucukelbir2017autodiffvi} or integrated Laplace approximation \citep{ruiz2012dynamicinla, Martino2019INLA}. For a surrogate model with non-identifiable parameters $\theta$, the algorithm would potentially have to deal with multimodalities \citep{pentwalk}.

\subsubsection{Second Step: Inference on Real Data (I-Step)}\label{subsubsection:second_step}
In the second step, real-world measurement data is given and our goal is to infer the unknown quantities of interest, i.e. the input simulation parameters, using the previously trained surrogate model as an efficient replacement of the complex simulator. 

\paragraph{Measurement Model}
We assume that the (implicit) real-world data generator is well described by the simulator $\cM$, but with a potentially different measurement noise $e_I$. Measurement data $y_I$ is then generated from an unknown underlying parameter $\omega_I$ (which has the same dimension as $\omega_T$ and serves as input to the simulator):
\begin{align}
    y_I = \cM(\omega_I; e_I) \quad \text{with} \quad e_I \sim p(e_I \mid \sigma_I),
\end{align}
where the measurement noise $e_I$ is drawn from a distribution $p(e_I \mid \sigma_I)$ with hyperparameters $\sigma_I$ that describe the aleatoric uncertainty of the measurement. This induces a generating distribution $p(y_I \mid \omega_I, \sigma_I)$ of measurements $y_I$ given inputs $\omega_I$ and noise hyperparameters $\sigma_I$. In contrast to the simulator setup, we only have access to the measurements $y_I$, but the underlying true input $\omega_I$ and $\sigma_I$ are unknown.

\paragraph{Inference}
Given a set of real-world measurement data $\ye = \{\ye^{(i)} \}^{N_I}_{i=1}$ with $\ye^{(i)} \sim p(y_I^{(i)} \mid \omega_I, \sigma_I)$ for $i = 1, \ldots, N_I$, our goal is to infer the posterior of the unknown parameters $\we$, which constitute our primary quantity of interest. Additionally, we can also infer $\sigma_I$ although it is only of secondary interest. To simplify the presentation, we drop $\sigma_I$ from the notation in the following.

If the simulator $\cM$ had a tractable and easy-to-evaluate likelihood function $p(y_I \mid \omega_I, \cM)$, we could simply calculate the posterior of the unknown simulation parameters $\omega_I$ given the measurement data $y_I$ and a prior $p(\omega_I)$:
\begin{align}
    p(\omega_I \mid y_I, \cM)\propto p(y_I \mid \omega_I, \cM)\, p(\omega_I).
\end{align}
However, for many real-world simulators, this likelihood is unavailable or highly cumbersome to evaluate (see Section~\ref{section:introduction}), which is why we seek to replace it with simpler likelihood of the previously trained surrogate $\widetilde{\cM}$.
To incorporate the uncertainties of the surrogate parameters $\theta = (c, \sigma_A)$ from the T-Step into our real-world inference, we need to \textit{somehow} propagate their T-Step posterior $p(\theta \mid \cD_T)$ to the I-step.
From the perspective of the I-Step, the uncertainty in $p(\theta \mid \cD_T)$ becomes aleatoric since it is no longer reducible. 
The main question how to propagate $p(\theta \mid \cD_T)$ turns out to have multiple answers -- even multiple ones fully justified by probability theory. Each of these answers corresponds to an uncertainty propagation method $u\in \cU$, for which we can obtain a surrogate-based posterior $p(\omega_I \mid y_I, \widetilde{\cM}, u)$ as detailed below.

\subsection{Uncertainty Propagation in the Two-Step Procedure}\label{subsection:up-2-step-procedure}
In the following, we present four different uncertainty propagation methods to perform the surrogate-based inference. These methods are namely (i) a Point Estimate, (ii) the Expected-Posterior (E-Post), (iii) the Expected-Likelihood (E-Lik), and (iv) the Expected-Log-Likelihood (E-Log-Lik). Accordingly, the set of considered propagation methods is given by $\cU =\{\text{Point}, \text{E-Post}, \text{E-Lik}, \text{E-Log-Lik}\}$. The three latter approaches propagate the uncertainty from the T-Step using the full posterior or samples from the posterior. In E-Post, we incorporate the T-Step posterior $p(\theta \mid \cD_T)$ directly into the posterior of the I-Step, while in the E-Lik and E-Log-Lik, the T-Step posterior is incorporated into the likelihood $p(\ye \mid \we, \widetilde{\cM}, u)$ of the I-step. In the following, all posteriors are calculated using the surrogate model $\widetilde{\cM}$ and we will omit the dependency of $\widetilde{\cM}$ in the posterior: $p(\omega_I \mid y_I, u) = p(\omega_I \mid y_I, \widetilde{\cM}, u)$ to improve readability. Furthermore, for all uncertainty propagation methods, we will assume that $\ye$ consists of conditional i.i.d. measurements $\ye^{(i)}$, such that the likelihood factorizes easily. This assumption is not necessary for our framework, but it simplifies the notation and allows for more efficient computation.

\subsubsection{Point Estimate}\label{subsubsection:Point}
First, we describe the I-Step using only a point estimator $\hat{\theta}$ of the T-posterior $p(\theta \mid \cD_T)$, e.g., the posterior mean, median, or mode. In this case, we condition the posterior of $\we$ on $\hat{\theta}$, i.e., we reduce the full posterior to a point estimate without epistemic uncertainty. This is not a method we advocate for, but rather use it as a baseline to compare against more sophisticated methods. The posterior probability of $\we$ given measurement data $\ye$ and $\hat{\theta}$ is then (up to normalizing constants):
\begin{align}
\label{eq:Point}
    p(\we \mid \ye, u= \text{Point} ) &\propto  p(\ye \mid \we, \hat{\theta})p(\we),
\end{align}
where the dependency of the posterior of $\we$ on $\hat{\theta}$ is represented via $u = \text{Point}$.
With the i.i.d. assumption, the (unnormalized) log I-posterior is
\begin{align}
\label{eq:Point-factorized}
\begin{split}
    \log p&(\we \mid \ye, u=\text{Point}) \\
     &\propto \sum_{i=1}^{N_I}\log p(\ye^{(i)} \mid \we, \hat{\theta}) + \log p(\we),
\end{split}
\end{align}
where we use the $\propto$ symbol for log-probability statements to imply that an additive constant $C$ (here, the log marginal likelihood) is not shown in the equation. Such a constant is independent of the parameters and thus irrelevant for posterior inference via MCMC or related sampling methods.
The log-posterior can easily be specified in a probabilistic programming language \citep{gorinova2019automatic} such as Stan \citep{carpenter2017stan} and we can then sample from the posterior with MCMC, leading to $K$ posterior samples $\{\we^{(1)}, \dots, \we^{(K)}\} \sim p(\we \mid \ye, u=\text{Point})$.
The benefit of the Point method is its comparably fast evaluation, since we only use a point estimate of the T-posterior and therefore can quickly evaluate the I-likelihood $p(y_I \mid \omega_I, \hat{\theta})$. While we are completely neglecting the epistemic uncertainty contained in the T-posterior, we propagate (a point estimate of) the surrogate approximation error parameter $\sigma_A$ through $\hat{\theta}$. We expect the Point I-posterior $p(\omega_I \mid y_I, u=\text{Point})$ to be overconfident (too narrow); unless we have a sufficiently large amount of simulation training data $\cD_T$ such that $p(\theta \mid \cD_T)$ converges to a point mass and then corresponds exactly to the point estimator $\hat{\theta}$.

\paragraph{Related work} Training a surrogate model using simulation data and subsequently employing its point estimate to infer unknown input parameters from observed data is a standard and well known approach \citep[e.g.,][]{marzouk2007stochasticsprectral, laloy2013effposteriorexpl, li2014adaptiveconstruction}. This method has been extensively studied in the context of Bayesian surrogate models, particularly through modularization introduced by \cite{bayarri2009modularizationbayesian}.

\subsubsection{Expected-Posterior}\label{subsubsection:E-Post}
Next, we present the Expected-Posterior (E-Post), a method that propagates both aleatoric and epistemic uncertainty in the T-posterior by marginalizing over the surrogate parameter posterior $p(\theta \mid \cD_T)$ in the posterior of $\omega_I$:
\begin{align}
\label{eq:E-post}
    p(\we \mid \ye, u=\text{E-Post}) =\int p(\omega_I \mid y_I, \theta) p(\theta \mid \cD_T)\mathrm{d}\theta.
\end{align}
Since we only have samples $\theta^{(s)}\sim p(\theta \mid \cD_T)$ from the T-step, we compute the Monte Carlo (MC) approximation of E-Post as:
\begin{align}
\label{E-post-mc}
    p(\we \mid \ye, u=\text{E-Post})\stackrel{\text{MC}}{\approx} \frac{1}{S}\sum_{s=1}^Sp(\omega_I \mid y_I, \theta^{(s)}),
\end{align}
which now is a finite mixture model with equal weights.
This can easily be implemented in a probabilistic program by fitting a separate model for each T-posterior draw $\theta^{(s)} \sim p(\theta \mid \cD_T)$ using the point I-posterior approach in \eqnref{eq:Point-factorized}, which results in $K$ draws of $\omega_I$, i.e., $\{\we^{(s, 1)}, \dots, \we^{(s, K)}\}$. The combination of these draws across all $s = 1, \ldots, S$ then represents an MC-estimate of the E-Post I-posterior as per \eqnref{E-post-mc}, with a total of  $S \cdot K$ posterior draws. We note that the number of propagated T-posterior draws $S$ is a hyperparameter that needs to be tuned depending on the complexity and dimensionality of the problem.

\paragraph{Related work} The idea of constructing a posterior via the aggregation of multiple posterior distributions, each approximated via samples, has been explored in multiple places in the literature. In the BayesBag method \citep{wadell2002bayesbagphylo, douady2003bayesbagmolecular, buhlmann2014bayesbag, huggins2020robust}, posteriors are obtained from bootstrapped copies \citep{efron1979bootstrap, breiman2004bagging} of the original dataset and subsequently averaging the resulting bootstrapped posteriors.
Similarly, in the context of missing value imputation \citep{little2019statistical}, this approach has been used to combine models fitted on multiple imputed datasets \citep{buerkner2017brms, buerkner2018advanced}. In terms of how many imputed data sets are needed, \cite{Austin2021missingdatainclinical} report that between 20 and 100 imputations are typically used. In both multiple data imputation and bagging, models are fitted to different datasets whereas in our E-Post method, the data is the same but the model itself changes. Furthermore, within the context of modularization of Bayesian models \citep{plummer2014cuts, jacob2017better}, this approach is known as the cut distribution.

\subsubsection{Expected-Likelihood}\label{subsubsection:E-Lik}
Next, we present the Expected-Likelihood (E-Lik) approach, where we marginalize over the T-posterior $p(\theta \mid \cD_T)$ in the I-likelihood:
\begin{align}
    p(\ye \mid \we, u=\text{E-Lik}) = \int p(\ye \mid \we, \theta)p(\theta \mid \cD_T) \mathrm{d}\theta.
\end{align}
The full I-posterior of E-Lik is then given by
\begin{align}
    p(\we \mid \ye, u=\text{E-Lik}) &\propto p(\ye \mid \we, u=\text{E-Lik}) p(\we).
\end{align}
Assuming a factorizable likelihood, we compute the log I-posterior over the whole dataset $\ye$ (without normalizing constant) as:
\begin{align}
\begin{split}
    &\log p(\we \mid \ye, u=\text{E-Lik}) \\
    &\propto \log \int \prod_{i=1}^{N_I}p(\ye^{(i)} \mid \we, \theta)p(\theta \mid \cD_T)\mathrm{d}\theta+\log p(\we),
\end{split}
\end{align}
for which we can obtain an MC approximation using draws $\theta^{(s)}\sim p(\theta \mid \cD_T)$:
\begin{align}
\begin{split}
    &\log p(\we \mid \ye, u=\text{E-Lik}) \\
    &\stackrel{\text{MC}}{\approx}\log \left( \frac{1}{S}\sum_{s=1}^{S} \prod_{i=1}^{N_I} p(\ye^{(i)} \mid \we, \theta^{(s)})\right) + \log p(\we).
\end{split}
\end{align}
This representation has the problem that the product over likelihood components becomes numerically unstable as $N_I$ grows larger, since the log operator cannot simply be pulled into sum over draws. To circumvent this, we use the \texttt{log-sum-exp} trick, i.e. calculate $\log \sum_{s=1}^S p(y_I \mid \omega_I, \theta^{(s)}) = \log \sum_{s=1}^S \exp(\log p(y_I \mid \omega_I, \theta^{(s)}))$, which has a numerically stable implementation \citep{carpenter2017stan}. For given $\theta^{(s)}$, the joint log likelihood is then again a simple sum: $\log p(y_I \mid \omega_I, \theta^{(s)}) = \sum_{i=1}^{N_I} \log p(y_I^{(i)} \mid \omega_I, \theta^{(s)})$.
In contrast to E-Post, E-Lik is expressed as a single probabilistic program and we can use MCMC to approximate its I-posterior with $K$ samples $\{\we^{(1)}, \dots, \we^{(K)}\}$.

More general, in contrast to E-Post, which marginalizes over $\theta$ in the I-posterior $p(\omega_I \mid y_I, \theta)$, the E-Lik method marginalizes over $\theta$ in the I-likelihood $p(y_I \mid \omega, \theta)$. E-Lik and E-Post are not identical 
(see Appendix \secref{subsection:counterexample} for a counterexample), 
% (see the Supplement \citep{reiser2023suppl} Section S2.2)
but we demonstrate in \secref{section:experiments} that they usually yield very similar results.
\paragraph{Related work}
To our knowledge, E-Lik constitutes a novel method for full uncertainty propagation. The approach used in \cite{Zhang2020WRR} appears related, although the details of their method are insufficiently described in the paper for a definitive assessment. Based on our understanding, their method can be seen as a special case of E-Lik, where the T-posterior is assumed to be normal and surrogate approximation error is ignored. Further, E-Lik is related to important quantities outside the area of UP: In particular, the log-likelihood constructed in E-Lik resembles the expected log predictive density (ELPD), a popular measure for predictive performance, which integrates the likelihood over the posterior of the same model before taking the logarithm outside the expectation \citep{vehtari2012bayespredmeth, vehtari2017loo, buerkner2022models}.
What is more, in the context of meta-analysis, \cite{blomstedt2019metaanalysis} proposed a method to combine posteriors resulting from different studies. While their sources of uncertainty are different than in our case, they also integrate them with an expected likelihood approach, rendering at least the core idea related to E-Lik.

\subsubsection{Expected-Log-Likelihood}\label{subsubsection:E-Log-Lik}
In the Expected-Log-Likelihood (E-Log-Lik) approach, instead of marginalizing over the likelihood as in E-Lik, we marginalize over the T-posterior in the \textit{log}-likelihood. We define the E-Log-Lik I-likelihood as:
\begin{align}
\begin{split}
    &p(\ye \mid \we, u=\text{E-Log-Lik}) \\
    &\vcentcolon = \exp\left(\int \log (p(\ye \mid \we, \theta)) \, p(\theta \mid \cD_T) \mathrm{d}\theta\right).
\end{split}
\end{align}
The posterior of the E-Log-Lik for $\we$ is then defined as:
\begin{align}
\begin{split}
    &p(\we \mid \ye, u=\text{E-Log-Lik}) \\
    &\propto p(\ye \mid \we, u=\text{E-Log-Lik}) p(\we).
\end{split}
\end{align}
When assuming a factorizable likelihood and ignoring the normalizing constant, the log posterior becomes
\begin{align}
\begin{split}
    &\log p(\we \mid \ye, u=\text{E-Log-Lik}) \\
    &\propto \int \log\left(\prod_{i=1}^{N_I}p(\ye^{(i)} \mid \we, \theta)\right)p(\theta \mid \cD_T) \mathrm{d}\theta + \log p(\we)\\
    &= \int \sum_{i=1}^{N_I}\log p(\ye^{(i)} \mid \we, \theta) \, p(\theta \mid \cD_T) \mathrm{d}\theta + \log p(\we),
\end{split}
\end{align}
The integral is readily approximated via draws from the T-posterior:
\begin{align}\label{eq:E-Log-Lik_mc}
\begin{split}
    &\log p(\we \mid \ye, u=\text{E-Log-Lik})\\
    &\stackrel{\text{MC}}{\approx} \frac{1}{S}\sum_{s=1}^S\sum_{i=1}^{N_I} \log p(\ye^{(i)} \mid \we, \theta^{(s)}) + \log p(\we),
\end{split}
\end{align}
which can be fitted with MCMC leading to $K$ draws $\{\we^{(1)}, \dots, \we^{(K)}\}$ in the I step.
We can further rewrite the MC approximation of the log-likelihood as
\begin{align}
\begin{split}
&\log p(y_I \mid \omega_I, u=\text{E-Log-Lik}) \\
&= \sum_{s=1}^S\sum_{i=1}^{N_I} \log \left( p(\ye^{(i)} \mid \we, \theta^{(s)})^{\frac{1}{S}} \right).
\end{split}
\end{align}
This shows that the E-Log-Lik can be interpreted as power-scaling the likelihood components with equal weights $1/S$ \citep{Geyer1991MarkovCM, kallioinen2022detecting},
which readily generalizes to unequal weights as we illustrate in \secref{subsubsection:clustering}.
However, in contrast to E-Lik and E-Post, this does not strictly follow rules of probability theory as we integrate over a probability measure in the log space.
Alternatively to E-Log-Lik, we could set up an Expected Log Posterior (E-Log-Post) as
\begin{align}
\begin{split}
    &p(\omega_I \mid y_I, u=\text{E-Log-Post})\\
    &\propto \exp\left(\int \log p(\omega_I \mid y_I, \theta) \, p(\theta \mid \cD_T)\mathrm{d}\theta\right),
\end{split}
\end{align}
% for arxiv: 
which however is equivalent to E-Log-Lik as we show in Appendix \ref{subsection:e-log-post}.
% which however is equivalent to E-Log-Lik as we show in S2.2.

\paragraph{Related work}
The computation of the E-Log-Lik is conceptually similar to a single E-Step in an Expectation-Maximization (EM) algorithm \citep{dempster1977em}, where the log likelihood is integrated over a discretized space of latent variable values (instead of T-posterior draws as is done here). Furthermore, the Gibbs loss, a computational convenient although uncommon measure of predictive performance, also calculates an expectation over the log-likelihood \citep{vehtari2012bayespredmeth, buerkner2022models}.

\subsubsection{Clustering of the T-Posterior Draws}\label{subsubsection:clustering}
To speed up computation of the I-Step while minimizing loss of information, we can reduce the number of propagated T-posterior draws via clustering (see \cite{Piironen2020projpred} for a related use case in the context of variable selection). 
For this purpose, any clustering algorithm can in principle be used, for example, KMeans clustering \citep{macqueen1967some, lloyd1982least}.
We apply the clustering algorithm to the set of T-posterior draws of the surrogate parameters $\{\theta^{(s)}\}_{s=1}^S$ to get cluster centroids $\{\mu^{(l)}\}_{l=1}^L$. The sufficient number of clusters $L$ for a trustworthy approximation of the I-posterior highly depends on the complexity of the I-posterior and is a hyperparameter that needs to be tuned. Here, we carried out a visual convergence analysis, given that rigorous guidelines on the choice of the number of clusters are lacking to date and an open scientific challenge. For each of the cluster centroids we additionally store weights $\{\alpha^{(l)}\}_{l=1}^L$, where $\alpha^{(l)}$ is the percentage of draws associated with the cluster. While inducing an approximation error, the cluster centroids together with the corresponding weights allow for a reliable and computationally efficient processing of the T-posterior draws as we show in \secref{section:experiments}.

Clustering and re-weighting can be easily applied in all UP methods by replacing $\theta^{(s)}$ with $\mu^{(l)}$ as well as replacing the equal weight $1/S$ with $\alpha^{(l)}$ after moving it inside the sum. For example, when using E-Log-Lik, the MC-approximated integral becomes:
\begin{align}
\begin{split}
    &\log p(y_I \mid \omega_I, u=\text{E-Log-Lik}) \\
    &\stackrel{\text{MC}}{\approx} \frac{1}{S}\sum_{s=1}^S\sum_{i=1}^{N_I} \log(p(\ye^{(i)} \mid \we, \theta^{(s)}) \\
    &\approx \sum_{l=1}^L \sum_{i=1}^{N_I}\alpha^{(l)} \log p(\ye^{(i)} \mid \we, \mu^{(l)}).
\end{split}
\end{align}

\subsubsection{Parallelization} 
\label{parallelization}
Inference based on all introduced UP methods can be parallelized, but to a different degree. To compute the I-posterior using E-Post, we fit a separate model for each T-posterior draw (or each cluster of draws). This is embarrassingly parallelizable, as the models are independent so can simply be run on different cores. However, for each model, a separate MCMC warmup phase is required, which induces computational overhead. 

In contrast, for E-Lik, we fit only one model during the I-step, which loops over the T-posterior draws when evaluating its likelihood. This however defines a more complicated posterior, which is substantially slower to sample from compared to the individual E-Post models. Fitting the single E-Lik model can be sped up by between-chain parallization when running multiple MCMC chains (say, one per core), but then we again create overhead due to separate warmup phases per chain.
Alternatively, even when running a single chain, the E-Lik model can be parallelized via within-chain parallelization (aka threading), where the likelihood contributions of the T-posterior draws are evaluated in parallel. An overhead occurs due to variable passing and other non-parallelized model parts (e.g., the prior density evaluation). This leads to diminishing returns in terms of the number of cores used for threading. For more details on threading in Stan, see \cite{buerkner2022brms_threading}.

E-Log-Lik behaves as E-Lik in terms of parallelizability as they both fit only a single model during the I-step. That said, in our experiments, E-Log-Lik models sampled substantially faster than E-Lik models, presumably for two main reasons. First, E-Log-Lik does not require the use of \texttt{log-sum-exp} in order to obtain the joint log-likelihood, since we aggregate directly on the log-scale. This reduces the number of required operations within each MCMC step. Second, presumably, the geometry of the E-Log-Lik I-posterior is simpler than that of the E-Lik I-posterior, thus implying a more efficient exploration with MCMC for the former.

\subsection{Evaluation of the Two-Step Procedure}\label{subsection:evaluation}
Checking the calibration of uncertainty estimates is an important step to improve the trustworthiness of any inference algorithm. As such, it is a crucial aspect of the Bayesian workflow \citep{gelman2013bayesian}. In our setup, we are specifically interested in the uncertainty calibration of $p(\omega_I \mid y_I, \widetilde{\cM}, u)$, that is our I-posterior implied by the surrogate. Simulation-based Calibration (SBC) checking \citep{cook2006validationquantiles, talts2018sbc, modrak2022sbctest} is a current gold-standard approach to validate Bayesian computation, jointly testing the trinity of the simulator, the probabilistic program, and the posterior approximation algorithm, e.g., a sampling algorithm such as MCMC.

We extend SBC to the two-step procedure and use the notation of \cite{buerkner2022models}. For any quantile $q \in (0,1)$, let $U_q(\omega_I \mid y_I, \widetilde{\cM}, u)$ be any uncertainty region (e.g., the quantile-based credible intervals or highest density intervals) given by the posterior $p(\omega_I \mid y_I, \widetilde{\cM}, u)$ that depends on the surrogate $\widetilde{\cM}$ and the uncertainty propagation method $u$ (see \secref{subsection:2-step-procedure} and \ref{subsection:up-2-step-procedure}). If the generating distribution of the assumed surrogate $\widetilde{\cM}$ is equal to the true data-generating distribution induced by the simulator the following property holds simultaneously for all $q \in (0,1)$:
\begin{align}\label{eq:sbc_self_consistency}
\begin{split}    
    q = \iint& \nI[\omega_I^* \in U_q(\omega_I \mid y_I, \widetilde{\cM}, u)] \, p(y_I \mid \omega_I^*, \cM) \\
    &\times \,p(\omega_I^*) \, dy_I \, d\omega_I^*,
\end{split}
\end{align}
where we ignore $\sigma_I$ for simplicity and $\nI[\cdot]$ denotes the indicator function. This self-consistency property tests the correct coverage of the uncertainty region conditional on the input for every quantile. Practically, we check that the prior draws are uniformly distributed in the surrogate-based samples of $\omega_I$. We calculate the ranks as $r(\omega_I^*, \{\omega_I^{(1)}, \dots, \omega_I^{(K)}\}) = \sum_{k=1}^{K} \nI\left[\omega_I^* \leq \omega_I^{(k)}\right]$ and test them for uniformity using graphical tests as proposed in \cite{saeilynoja2022ecdf}. In addition to graphical tests, we calculate the $\log(\gamma)$-statistic \citep{saeilynoja2022ecdf} as a quantitative measure of uniformity allowing for a faster comparison of calibration (or strength of miscalibration) between different methods. Checking the calibration of a specific uncertainty region, e.g., $U_{0.95}(\omega_I \mid y_I, \widetilde{\cM}, u)$, corresponds to evaluating \eqnref{eq:sbc_self_consistency} for $q = 0.95$ and is a special case of SBC.

First, as a point of comparison, we explain standard SBC for inference using the simulator $\cM$: (i) Sample a simulation input parameter $\omega_I^* \sim p(\omega_I)$, (ii) conditioned on $\omega_I^*$ generate measurement output data $y_I \sim p(y_I \mid \omega_I^*)$, (iii) using the simulator $\cM$ and given the measurements $y_I$ draw samples $\{\omega_I^{(1)}, \dots, \omega_I^{(K)}\}$, and (iv) using the posterior samples, we calculate the rank statistics $r(\omega_I^*, \{\omega_I^{(1)}, \dots, \omega_I^{(K)}\})$ from the posterior $p(\omega_I \mid y_I, \cM)$. This procedure is repeated for a chosen number of I-Step trials and uniformity is tested on the stored ranks, as described above. 

However, in the two-step procedure (see \secref{subsection:2-step-procedure}), where we additionally train a surrogate given simulation data (T-Step) and then propagate its uncertainty to the I-Step, we need to extend SBC as shown in \algoref{alg:sbc}: We repeat the T-step multiple times (number of T-Step trials) and, in each iteration, we simulate a new training dataset $\cD_T$ used to train the surrogate $\widetilde{\cM}$. Within each such T-Step trial, we repeat the I-Step for SBC: First, we draw a sample of the simulation parameters $\omega_I^*$ and the simulation noise hyperparameter $\sigma_I$ from their respective priors. Next, measurement data $y_I$ is generated using the simulator $\cM$. Using the surrogate $\widetilde{\cM}$ and the UP method $u$, we draw posterior samples $\{\omega_I^{(1)}, \dots, \omega_I^{(K)}\}$ from the I-posterior and store the ranks $r(\omega_I^*, \{\omega_I^{(1)}, \dots, \omega_I^{(K)}\})$. The repetition of the T-Step and I-Step, in addition to covering the whole input space of $\omega_I$, helps to marginalize over both the noise of the simulator and the noise in the (assumed) measurement process.

With this SBC variant of the two-step procedure, we simultaneously test six different scenarios, where a failure can indicate one or more of the following scenarios: 
\begin{itemize}
    \item scenarios also tested in standard SBC
    \begin{enumerate}[label=(\roman*)]
        \item incorrect implementation of simulator $\cM$
        \item incorrect implementation of probabilistic program of the surrogate
        \item problems with the sampling algorithm
    \end{enumerate}
    \item additional scenarios in proposed SBC for surrogate-based inference
    \begin{enumerate}[label=(\roman*)]
    \setcounter{enumi}{3}
        \item inflexible surrogate
        \item insufficient training of surrogate because of too little simulation training data
        \item inappropriate uncertainty propagation in the surrogate-based inference.
    \end{enumerate}
\end{itemize}
Regarding points (i)-(iii), we assume the simulator to be correct and the surrogate to be relatively simple (implementation-wise) as well as easy to fit using MCMC. Concerning points (iv) and (v), we need a sufficiently flexible surrogate in order to remove the approximation error (i.e., $\sigma_A \to 0$)  and an infinite amount of training samples ($N_T \to \infty$) in order to remove the epistemic uncertainty in the posterior $p(\theta \mid \cD_T)$. However, practically the latter will not be the case and we expect the calibration to be imperfect. Nonetheless, we will still be able to compare surrogate models $\widetilde{\cM}$ and different UP methods $u$ by comparing their SBC results, either graphically or via test statistics.
\begin{algorithm*}
\caption{SBC for the Two-Step Procedure}\label{alg:sbc}
\begin{algorithmic}
\State Choose simulator $\cM$, surrogate $\widetilde{\cM}$, and uncertainty propagation method $u$
\For{$m$ in Number T-Step trials}
    \State Draw a training dataset $\cD_T$ using the simulator $\cM$
    \State Fit the surrogate $\widetilde{\cM}$ and calculate the posterior $p(\theta \mid \cD_T)$ (T-Step)
    \For{$n$ in Number I-Step trials}
        \State Draw a prior sample $\omega_I^* \sim p(\omega_I)$, $\sigma_I \sim p(\sigma_I)$
        \State Draw measurements $y_I \sim p(y_I \mid \omega_I^*, \sigma_I, \cM)$ using simulator $\cM$
        \State Draw posterior samples $\{\omega_I^{(1)}, \dots, \omega_I^{(K)}\} \sim p(\omega_I \mid y_I, \widetilde{\cM}, u)$ (I-Step)
        \State Store the rank of $\omega_I^*$ within the set of posterior samples $\{\omega_I^{(1)}, \dots, \omega_I^{(K)}\}$
    \EndFor
\EndFor
\State Perform uniformity test on the stored ranks
\end{algorithmic}
\end{algorithm*}

\section{Experiments}\label{section:experiments}
We evaluate our surrogate-based Bayesian inference framework in three case studies: (1) A linear setup, where we propagate only epistemic uncertainty, (2) a nonlinear setup, in which we propagate both epistemic and aleatoric uncertainty, and (3) a real-world model. All code and material can be found on GitHub\footnote{\url{https://github.com/philippreiser/bayesian-surrogate-uncertainty-paper}}.

\subsection{Case Study 1: Uncertainty Propagation in a Linear Model}\label{subsection:linear_model}

In the first case study, we use a linear setup leading to partly analytic posteriors that allow us to study our framework in a simple, well-understood scenario.

\paragraph{Setup}
We consider a simple linear model as simulator:
\begin{align}
    y_T = \cM(\omega_T, \sigma_S=0) = a + b \omega_T,
\end{align}
with simulation input parameter $\omega_T$ and two simulation control parameters $a, b$ set to $a = 0.5$, $b = 2$, and a simulation noise parameter $\sigma_S$ set to zero, i.e., we use a deterministic simulator.
For the T-Step, we consider $N_T=2$ training points $\cD_T = \{ \ws^{(i)}, \ys^{(i)} \}_{i=1}^{\Ns}$ (chosen according to \tabref{table:setup_linear_case_study}), where $\omega_T^{(i)}$ denotes the simulation input and $y_T^{(i)} = \cM(\omega_T^{(i)})$ the corresponding simulation output.  We denote by $\Omega_T = \bigl[ \begin{smallmatrix}1 & \omega_T^{(1)}\\ 1 &\omega_T^{(2)}\end{smallmatrix}\bigr]$ the design matrix of all input parameters and by $y_T$ the vector of all simulation outputs.

We set the surrogate model to a linear model as well, that is, we use the same model class as for the simulator:
\begin{align}
    \widetilde{\cM}(\omega_T, c) = c_1 + c_2\omega_T,
\end{align}
where the intercept $c_1$ and slope $c_2$ form the surrogate approximation parameters $c = [c_1, c_2]^\top$.
For the T-Step, we only consider the surrogate approximation parameters $c$ as trainable surrogate parameters. Even though the surrogate can match the simulator perfectly, we fix the surrogate approximation error hyperparameter $\sigma_A$ to values greater than zero. This allows us to control the width of the T-posterior and thus the epistemic uncertainty, as shown below. 
As prior on the surrogate approximation parameters we choose a bivariate normal distribution:
\begin{align}
    p(c) = \mathcal{N}(c \mid \mu_{T0}, \Sigma_{T0}),
\end{align}
with mean $\mu_{T0}=0$ and a covariance matrix $\Sigma_{T0} = \sigma_{T0}^2I$ where $I$ is the identity matrix.
We set the T-likelihood to a normal distribution as well:
\begin{align}
    p(y_T \mid c) = \prod_{i=1}^{N_T} \mathcal{N}(y_T^{(i)} \mid c_1 + c_2\omega_T^{(i)}, \sigma_A^2).
\end{align}
To generate data for the I-Step, a single measurement $y_I = \cM(\omega_I^*)$ is obtained by inputting a true input parameter $\omega_I^*$ to the deterministic simulator $\cM$. In the following, we will also use the augmented vector $\hat{\omega}_I=[1, \omega_I]^T$ to simplify notation.
Within the I-step, we set a normal prior with mean $\mu_{I0}$ and variance $\sigma_{I0}^2$ on $\omega_I$:
\begin{align}
    p(\omega_I) = \mathcal{N}(\omega_I \mid \mu_{I0}, \sigma_{I0}^2).
\end{align}
The I-likelihood $p(y_I \mid \omega_I)$ is assumed to be normal with fixed variance $\sigma_{I}^2$:
\begin{align}
    p(y_I \mid \omega_I, c) = \mathcal{N}(y_I \mid c_1 + c_2\omega_I, \sigma_{I}^2)
\end{align}

\paragraph{Deriving the posteriors} 
In the following, we perform the T-Step (see \secref{subsubsection:first_step}) and I-Step (see \secref{subsection:up-2-step-procedure}) for the linear setup. In the T-Step, we calculate the T-posterior of the surrogate approximation parameters $c$, which contains epistemic uncertainty. Then we derive the four different I-step methods when propagating only the epistemic uncertainty from the T-posterior. The aleatoric uncertainty is zero because, in our setup, the simulator is within the approximation space of the surrogate model.

\paragraph{T-Step}
We calculate the T-posterior for the surrogate approximation parameters $c$ given the training data $\cD_T$ from the simulator by using the conjugate prior relation for a normal-normal model \citep{murphy2007conjugate, pml1Book} leading to a normal posterior:
\begin{align}
    p(c \mid \cD_T) &=\mathcal{N}(c \mid \mu_{T1}, \Sigma_{T1}),
\end{align}
with
\begin{align}
    \Sigma_{T1} &= (\Sigma_{T0}^{-1} +\sigma_A^{-2}\Omega_T^\top\Omega_T)^{-1}, \\
    \mu_{T1} &= \Sigma_{T1}(\Sigma_{T0}^{-1}\mu_{T0} + \sigma_A^{-2}\Omega_T y_T).
\end{align}
We see that by fixing $\sigma_A$ we can control the width of the T-posterior and hence the epistemic uncertainty of the surrogate parameters, even if we do not propagate $\sigma_A$ itself.

\paragraph{I-Step}
Below, we derive the I-posteriors of all four uncertainty propagation procedures.
We compute the mean of the T-posterior: $\bar{c} = \mu_{T1}$ and use it to calculate the Point I-posterior as follows:
\begin{align}
    &p(\omega_I \mid y_I, u = \text{Point}) \propto  p(\omega_I)p(y_I \mid \omega_I, \bar{c})\\
    &= \mathcal{N}(\omega_I \mid \mu_{I0}, \sigma_{I0}^2)  \mathcal{N}(y_I \mid \mu_{T1}^{(1)} + \mu_{T1}^{(2)}\omega_I, \sigma_{I}^2) \\
    &\propto \cN(\we \mid  \mu_{I1}, \sigma_{I1}^2),
\end{align}
with
\begin{align}
        \sigma_{I1}^2 &= (\sigma_{I0}^{-2} +\se^{-2} \mu_{T1}^{(2)}\mu_{T1}^{(2)})^{-1}, \\
        \mu_{I1} &= \sigma_{I1}^2(\sigma_{I0}^{-2}\mu_{I0} + \se^{-2}\mu_{T1}^{(2)}(\ye-\mu_{T1}^{(1)})).
\end{align}
Again, this works due to the normal-normal conjugacy. We see that the I-posterior variance $\sigma_{I1}^2$ does not depend on the T-posterior covariance matrix $\Sigma_{T1}$, i.e. the uncertainty from the surrogate training step is neglected.

The E-Log-Lik I-posterior is given by
\begin{align}
\begin{split}
    &p(\we \mid \ye, u=\text{E-Log-Lik}) \\
    &\propto p(\we)\exp\left\{\int \log(p(\ye \mid \we, c))p(c \mid \cD_T) \mathrm{d}c\right\} \\
    &= \mathcal{N}(\omega_I \mid \mu_{I0}, \sigma_{I0}^2)\\
    &\times \exp\left\{\int \log(\mathcal{N}(y_I \mid \hat{\omega}_I^\top c, \sigma_{I}^2))\mathcal{N}(c \mid \mu_{T1}, \Sigma_{T1})\mathrm{d}c\right\} \\
    &\propto \cN(\omega_I \mid  \mu_{I1}, \sigma_{I1}^2),
\end{split}
\end{align}
with
\begin{align}\label{eq:e_log_lik_variance}
\begin{split}
        \sigma_{I1}^2 =& (\sigma_{I0}^{-2} +\se^{-2} (\mu_{T1}^{(2)}\mu_{T1}^{(2)} + \Sigma_{T1}^{(2,2)}))^{-1}, \\
        \mu_{I1} =& \sigma_{I1}^2(\sigma_{I0}^{-2}\mu_{I0} \\
        &+ \se^{-2}(\mu_{T1}^{(2)}\ye-\Sigma_{T1}^{(1,2)}-\mu_{T1}^{(1)}\mu_{T1}^{(2)}))
\end{split}
\end{align}
Accordingly, it is also a normal distribution, but a different one from the Point I-posterior. 
The detailed derivation of the E-Log-Lik is given in Appendix \ref{subsubsection:derivation_e-log-lik}.
% The detailed derivation of the E-Log-Lik is given in S2.3.
If we look at the variance $\sigma_{I1}^2$  we see that E-Log-Lik produces counter-intuitive results, as $\Sigma_{T1}^{(2,2)}$ and $\sigma_{I1}^2$ are reciprocally related. That is, as the surrogate model gets more uncertain in the T-step, the I-posterior gets more certain.

The E-Lik I-posterior is computed as
\begin{align}
\begin{split}
    p(\we& \mid \ye, u=\text{E-Lik}) \\
    \propto \; & p(\we) \int p(\ye \mid \we, c) p(c \mid \cD_T) \mathrm{d}c\\
    = \; & \mathcal{N}(\omega_I \mid \mu_{I0}, \sigma_{I0}^2)\\
    &\times \int \mathcal{N}(y_I \mid  \hat{\omega}_I^\top c, \sigma_I^2)\mathcal{N}(c \mid \mu_{T1}, \Sigma_{T1})\mathrm{d}c \\
    = \; & \mathcal{N}(\omega_I \mid \mu_{I0}, \sigma_{I0}^2)\\
    &\times \mathcal{N}(y_I \mid \hat{\omega}_I^\top  \mu_{T1}, \hat{\omega}_I^\top\Sigma_{T1}\hat{\omega}_I+\sigma_{I}^2).
\end{split}
\end{align}
Here, we cannot apply the normal-normal conjugacy, because $\hat{\omega}_I$ is present in both the mean and the variance in the second term of the product of the two normals. Instead, the resulting distribution is non-analytic and we have to use numerical integration to calculate the normalization constant. Looking at the derived quantity, we see that the variance of the term resulting from the marginalization over the surrogate approximation parameters $c$ increases with the variance of the T-posterior $\Sigma_{T1}$. 

Finally, we calculate the I-posterior using the E-Post approach:
\begin{align}
\begin{split}
    p(\we& \mid \ye, u=\text{E-Post}) \\
    = &p(\omega_I) \int \frac{p(y_I \mid \omega_I, c)p(c \mid \cD_T)}{\int p(y_I \mid  \omega_I, c) p(\omega_I) \mathrm{d}d\omega_I}  \mathrm{d}c \\
    = &\mathcal{N}(\omega_I \mid \mu_{I0}, \sigma_{I0}^2) \\
    &\times\int \frac{\mathcal{N}(y_I \mid \hat{\omega}_I^\top c, \sigma_I^2) \mathcal{N}(c \mid \mu_{T1}, \Sigma_{T1})}{\mathcal{N}(y_I \mid c_1 + c_2\mu_{I0}, c^\top c \sigma_{I0}^2+\sigma_{I}^2)} \mathrm{d}c.
\end{split}
\end{align}
Again, this integral is non-analytic and numerical integration is required to calculate the E-Post I-posterior. Nevertheless, we can expect similar behavior to E-Lik, since we also marginalize over the T-posterior and only normalize differently.

\paragraph{Results}
\begin{table*}
  \centering
  \caption{Parameters with realized values for case study 1.}
  \renewcommand{\arraystretch}{1.2}
  \begin{tabular}{|p{2cm}|c|c|c|c|c|c|c|c|c|c|}
    \hline
    \multirow{2}{2cm}{\textbf{Parameters}} & \multicolumn{5}{c|}{\textbf{T-Step}} & \multicolumn{4}{c|}{\textbf{I-Step}} \\
    \cline{2-10}
    & $\begin{bmatrix} \omega_T^{(1)} \\ \omega_T^{(2)}\end{bmatrix}$ & $\begin{bmatrix} a \\ b \end{bmatrix}$ & $\mu_{T0}$ & $\sigma_{T0}$ & $\sigma_A$ & $\omega_I^*$ & $\mu_{I0}$ & $\sigma_{I0}$& $\sigma_I$ \\
    \hline
    \textbf{Values} & $\begin{bmatrix} -0.9 \\ -0.3 \end{bmatrix}$ & $\begin{bmatrix}0.5 \\ 2 \end{bmatrix}$ & $\begin{bmatrix} 0 \\ 0 \end{bmatrix}$ & $10$ & $\{0.1, 0.5, 1\}$ & -0.5 & 0 & 1 & 0.1 \\ \hline
  \end{tabular}\
  \label{table:setup_linear_case_study}
\end{table*}

\begin{figure*}
    \begin{center}
    \includegraphics[width=\textwidth]{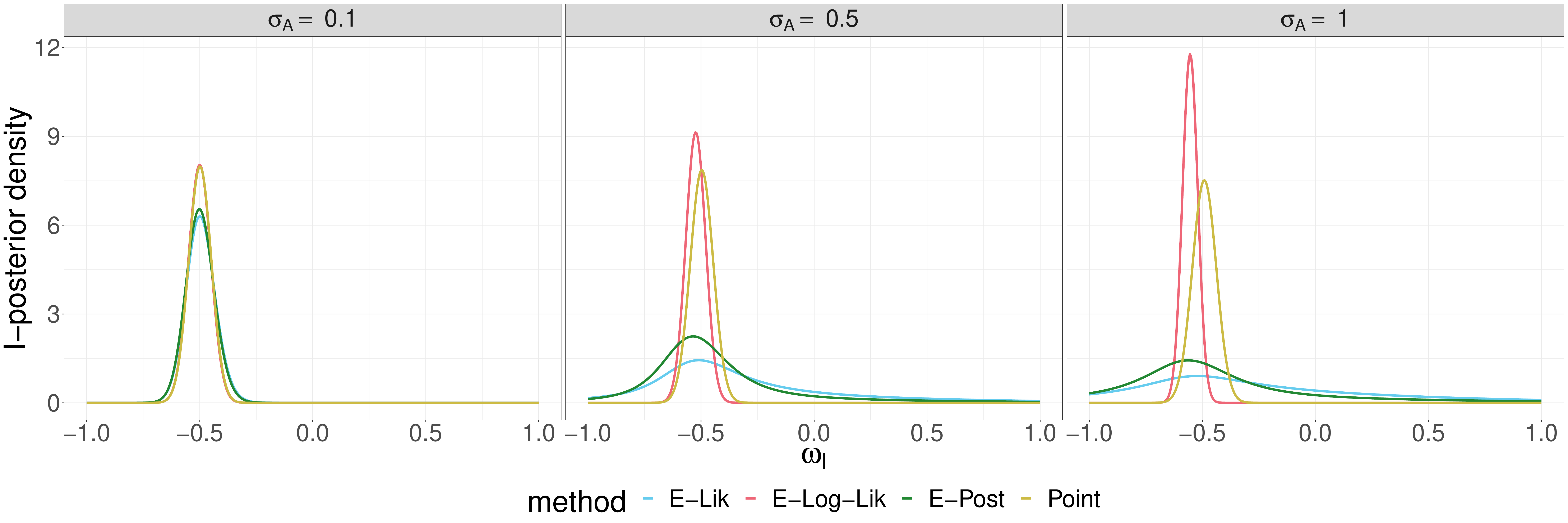}
    \caption{I-posterior densities for the linear surrogate with normal priors/likelihoods in case study 1. We use the data and parameters as specified in \tabref{table:setup_linear_case_study}. We use four different UP methods to compute the I-posterior while the surrogate approximation error $\sigma_A = \{0.1, 0.5, 1\}$ is varied.}
    \label{fig:analytic_istep_posteriors}
    \end{center}
\end{figure*}

In \tabref{table:setup_linear_case_study}, we provide an overview over the chosen training data, measurement data, and hyperparameters for the T and I-Step. We specifically vary the surrogate approximation error parameter $\sigma_A = \{0.1, 0.5, 1\}$ to control the epistemic uncertainty during the surrogate training, as explained above. To compare the four I-Steps, we calculate their I-posteriors under the given scenarios. The results are illustrated in \figref{fig:analytic_istep_posteriors}. 
We see that Point produces results that are independent of the standard deviation of the T-likelihood. This is expected since it does not propagate the T-epistemic uncertainty at all.
Intuitively, for the other three methods, the I-posteriors should become more uncertain as we increase the uncertainty in the T-step, since our surrogate model gets less trustworthy. This is indeed what happens for E-Lik and E-Post, which also produce very similar but not identical results. In contrast, E-Log-Lik behaves counter-intuitively since its I-posterior becomes more \textit{certain} as the T-posterior becomes more uncertain, a behavior that we also see clearly from \eqnref{eq:e_log_lik_variance}, as noted above.

\subsection{Case Study 2: Uncertainty Propagation in a Logistic Model}\label{subsection:case_study_2}
The second case study examines a nonlinear problem where we have to rely on MCMC for the posterior approximations, because analytic posteriors are unavailable.
We choose the one-dimensional logistic function
\begin{align}
    y = \cM(\omega) = \frac{2}{1+\exp(-10 \omega)}-1
\end{align}
as the (true) simulator since it is invertible and smooth everywhere. We consider two different surrogate models: The first is a parameterized generalization of the simulator and the second is a polynomial chaos expansion (PCE) surrogate. We will discuss these two cases separately below.

\subsubsection{Logistic Surrogate Model}\label{subsubsection:logistic_true}
\paragraph{Setup}
As surrogate, we consider \begin{align}
    \widetilde{\cM}(\omega; c) = \frac{\alpha}{1+\exp(-\beta (\omega-\gamma))}+\delta,
\end{align} 
with surrogate approximation parameters $c = (\alpha, \beta, \gamma, \delta)$. Here, the true simulator is contained in the set of surrogate models (for $\alpha = 2, \beta = 10, \gamma = 0, \delta = -1$).

For the T-Step, we generate the training set by setting the input points $\omega_T^{(i)} \in [-1, 1]$ to the first $N_T$ points of a slightly modified one-dimensional Halton sequence \citep{halton1960quasirandomseq}, which starts with the boundary points ($-1$ and  $1$) and then progresses as the standard Halton sequence with center point 0.
The simulated output responses $\ys^{(i)}$ for $i \in \{1, \dots, \Ns\}$ are sampled from a normal distribution with mean equal to the evaluated logistic simulator $\cM$ at the input points and standard deviation $\sigma_S$, i.e. $\ys^{(i)} \sim \cN(\cM(\ws^{(i)}), \sigma_S^2)$. 
To avoid sampling issues during the training step, we induce a small simulation noise $\sigma_S=0.01$, which is not explicitly modelled.

For the surrogate parameters $c$, we specify normal priors with means around the true values and a standard deviation of 1 (except for $\beta$, where we set the standard deviation to 10). For the T-likelihood, we consider a normal distribution:
\begin{align}
    p(y_T \mid c) = \cN(y_T \mid \widetilde{\cM}(\omega_T, c), \sigma_A^2)
\end{align}
with the surrogate approximation hyperparameter $\sigma_A$.
On each training data set, we fit the parameters of our surrogate using Markov chain Monte Carlo (MCMC). Specifically, we use the no-U-turn-sampler (NUTS) \citep{hoffman2014nuts}, an adaptive form of Hamiltonian Monte Carlo (HMC) \citep{neal2011} which is a gradient based MCMC sampler via the probabilistic programming language Stan \citep{standev2024stan}.
We use four chains, each running for 1250 iterations (1000 warmup and 250 post-warmup iterations), resulting in a total of 1000 T-posterior draws. We assessed convergence using standard convergence checks (i.e., the R-hat diagnostic of all parameters \citep{vehtari_rhat_2021}).

For the I-Step, we generate $\Ne = 5$ random measurements $\ye \sim \cN(\cM(\we^*), \se^2)$, based on the simulator, true input parameters $\we^*$, and the measurement error $\se=0.01$. As priors we set $\we \sim \cN_{[-1, 1]}(0, 0.5^2)$ (with truncation bounds $[-1, 1]$) and $\se \sim \text{uniform}(0, 0.05)$. These hyperparameters were chosen so that the true simulator could make valid inference about the input parameters given the measurement data.

In contrast to case study 1, we propagate the T-posterior through samples and hence use the MC approximation (see \secref{subsection:up-2-step-procedure}) for each of the four methods (Point, E-Lik, E-Log-Lik, E-Post). We sample from the I-posterior of $\we$ and $\se$ using NUTS with four chains, each running for 5000 iterations (1000 warmup and 4000 post-warmup iterations), resulting in a total of 16000 I-posterior draws.

Similar to case study 1, we propagate only the epistemic uncertainty that is encoded in the T-posterior of the surrogate parameters $c$. For this purpose, we either utilize all T-posterior draws or employ KMeans clustering (see \secref{subsubsection:clustering}) with $L=25$ clusters. As the simulator is contained in the class of surrogates, no approximation error is present ($\sigma_A = 0$) and hence, there is no aleatoric uncertainty to propagate. 

\begin{figure*}
    \begin{center}
    \includegraphics[width=\textwidth]{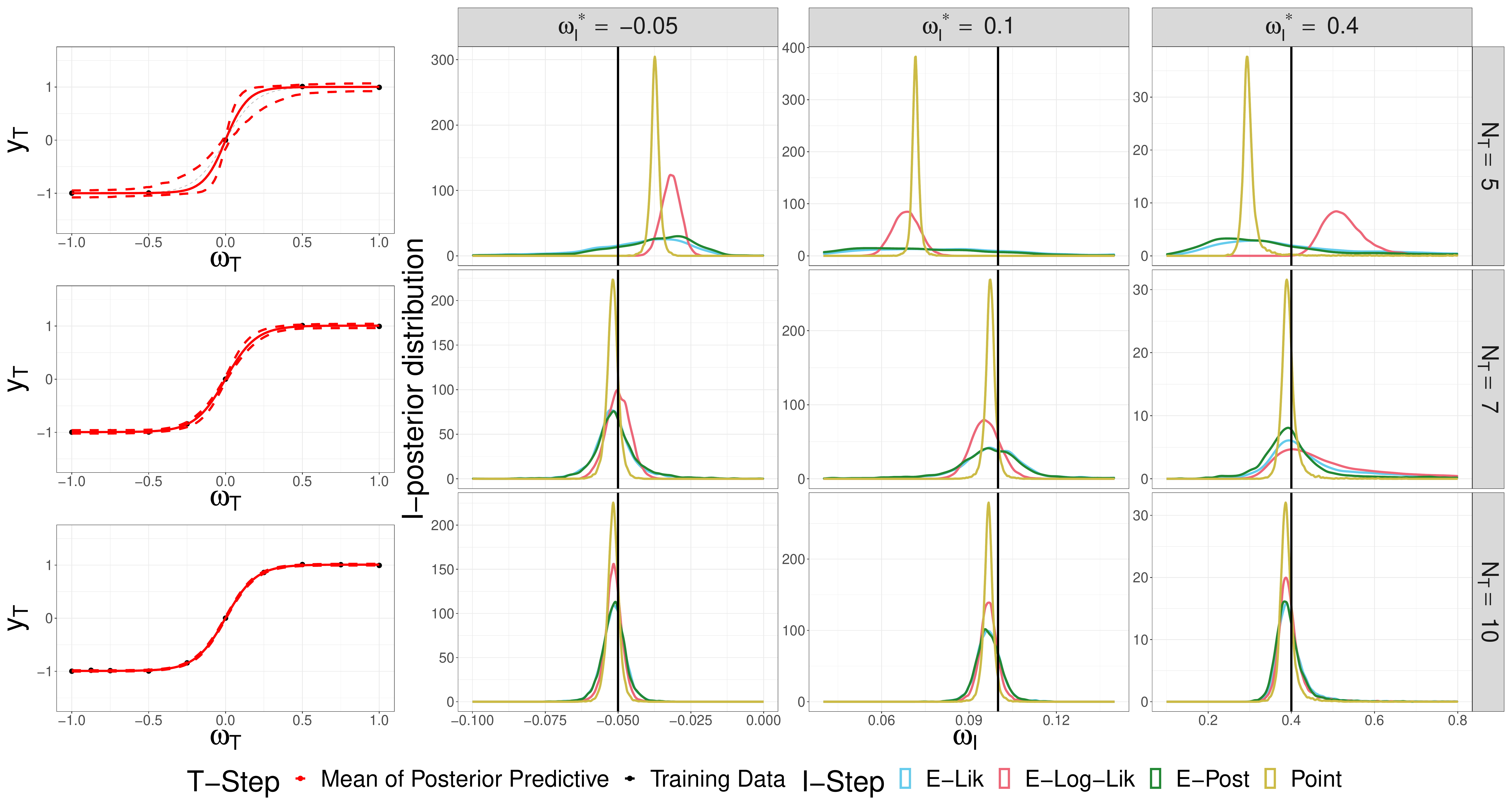}
    \caption{Selected results for two-step procedure with the logistic surrogate in case study 2. Left: For $\Ns = \{5, 7, 10\}$ the training data set $\cD_T$ (black dots) and the mean of the T-posterior predictive distribution (red lines) is shown. Right: For each underlying true input $\we^* \in \{-0.05, 0.1, 0.3\}$ (black vertical lines), we depict the I-posterior distributions for each Point, E-Lik, E-Post, and E-Log-Lik (colored lines).}
    \label{fig:i_step_different_s_steps}
    \end{center}
\end{figure*}

\paragraph{Posterior Distributions}
We set $\sigma_S = 0.01$ and vary the number of training points $N_T = \{5, 7, 10\}$ to perform the T-Step. In \figref{fig:i_step_different_s_steps}, the left column shows the mean of the T-posterior predictive with the 95\%-credible interval (CI) to display its epistemic uncertainty. As expected, increasing $N_T$ leads to smaller T-epistemic uncertainty. In Appendix \figref{fig:pairs_plot_t_posterior} we depict the pairs plot of the T-posterior draws of the logistic surrogate.

We compare the I-posteriors resulting from the four different methods using the measurements resulting from exemplary underlying true inputs $\we^* \in \{-0.05, 0.1, 0.3\}$ in the three right columns in \figref{fig:i_step_different_s_steps}.
The Point method yields I-posteriors with constant width regardless of variation in $N_T$. The I-posteriors of E-Lik and E-Post show similar behavior and become more uncertain as T-epistemic uncertainty increases. The E-Log-Lik follows a similar trend, but its I-posteriors have qualitatively different shapes and are narrower. 
For $N_T=5$, only E-Lik and E-Post have substantial I-posterior probability mass on the true inputs $\omega_I^*$. Notably, all uncertainty propagation methods converge to the same results as the epistemic uncertainty from the T-Step decreases.

\paragraph{Calibration}
To check if the methods for estimating the I-posteriors are calibrated, we use SBC checking \citep{talts2018sbc, modrak2022sbctest} with the \texttt{SBC} R package \citep{shinyoung2023sbcpackage}. Concretely, we perform the adapted SBC procedure for surrogate-based inference as detailed in \secref{subsection:evaluation}. For the I-Step trials, we simulated the true values of the inputs $\omega_I^*$ and measurement error $\sigma_I$ from the above chosen priors. We perform 20 I-step trials within each of 10 T-Step trials resulting in a total of 200 SBC-trials. 

\begin{figure*}
    \begin{center}
    \includegraphics[width=\textwidth]{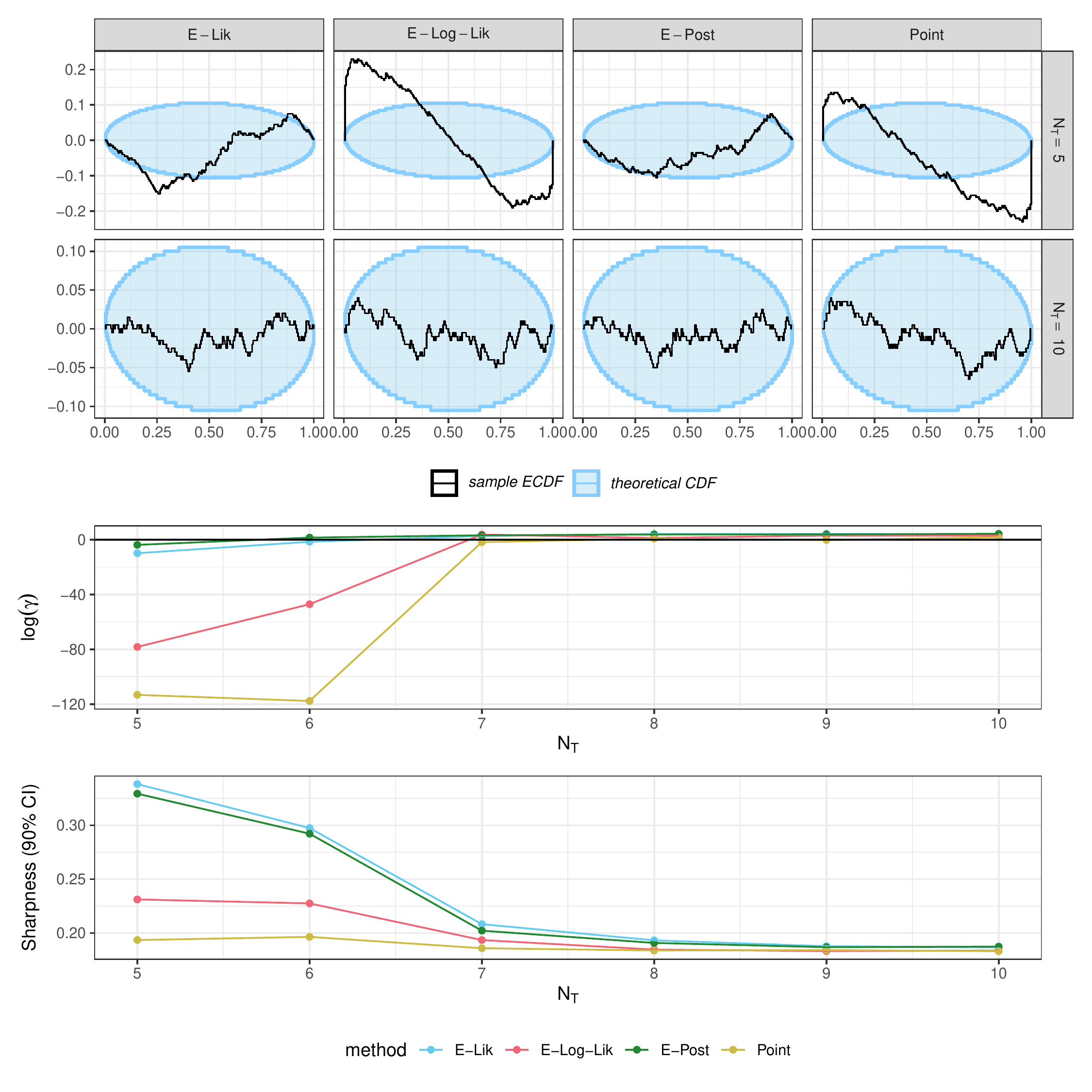}
    \caption{Calibration and sharpness of the I-posteriors using logistic surrogate in case study 2. Top: ECDF difference plots for the I-posterior distributions of $\omega_I$ resulting from the four different methods. The blue areas in the ECDF difference plots indicate 95\%-confidence envelopes and the black lines indicate the empirical cumulative distribution function (ECDF) for two different number of simulation points $N_T = \{5, 10\}$. Center: log-gamma-statistics of SBC with calibration threshold depicted as black horizontal line. Bottom: sharpness (90\% CI) of I-posterior for four different I-Steps (colored dots/lines) for $N_T = \{5, 6, 7, 8, 9, 10\}$.}
    \label{fig:sbc_ecdf}
    \end{center}
\end{figure*}

To evaluate the calibration of the four methods graphically, we show the empirical cumulative distribution function (ECDF) difference plots  \citep{saeilynoja2022ecdf} in the upper part of \figref{fig:sbc_ecdf}. We choose two scenarios: high and low T-epistemic uncertainty, represented by the number of training points $N_T = \{5, 10\}$. In this graphical test, calibration is achieved if the black line lies within the blue region, i.e. the 95\%-confidence envelops. We observe that, while high T-epistemic uncertainty is present, E-Post and E-Lik are almost calibrated, whereas Point and E-Log-Lik are overconfident. For $N_T = 8$ all methods show good calibration.

As an additional continuous calibration metric, we calculate the $\log(\gamma)$-statistic \citep{modrak2022sbctest} from our SBC results. We vary $\Ns = \{5, 6, 7, 8, 9, 10\}$ to control for the amount of to-be-propagated T-epistemic uncertainty. The center part of \figref{fig:sbc_ecdf} shows the corresponding results. As a general trend, we observe that E-Lik and E-Post are similarly calibrated. While first ($N_T=5$) slightly miscalibrated, for $N_T>5$ proper calibration is achieved. In contrast, E-Log-Lik and Point are both miscalibrated for $N_T<7$, but with more training data, as the T-epistemic uncertainty diminishes, they also become well calibrated.

In the bottom of \figref{fig:sbc_ecdf} we depict the sharpness \citep{gneiting2007probforecastsharpness, buerkner2022models} of the I-posteriors, here measured by the width of the 90 \% CI of $\omega_I$. For similarly calibrated methods, we say that the method with the smaller CI is sharper. We observe that Point and E-Log-Lik produce overall sharper results, but given their bad calibration, its clear that these two methods are just overconfident.

\subsubsection{Polynomial Surrogate Model}\label{subsubsection:logistic_pce}
\paragraph{Setup}
We now make the task substantially harder by not including the true simulator in the set of considered surrogate models. For this purpose, we use a polynomial chaos expansion (PCE) model as surrogate \citep{Wiener1938JHUP, Sudret2008RESS, Oladyshkin2019RESS, Buerkner2022bspce}:
\begin{align}
    \widetilde{\cM}(\omega; c) = \sum_{i=0}^d c_i \psi_i(\omega),
\end{align}
with the vector of surrogate coefficients $c = (c_0, ..., c_d)$, the maximum degree of polynomials $d$, and Legendre polynomials $\psi_i(\omega)$ (see \cite{Sudret2008RESS} for a detailed definition). We consider wide, independent normal priors for all surrogate coefficients: $c_i \sim \cN(0, 5)$. In the following, we fix the maximum polynomial degree to $d=5$. By doing so, we create a scenario in which our surrogate is unable to fit the underlying true model appropriately such that an approximation error $e_A$ is induced. This creates a scenario in which it is important to propagate both T-epistemic and T-aleatoric uncertainty.

\begin{figure*}
    \begin{center}
    \includegraphics[width=\textwidth]{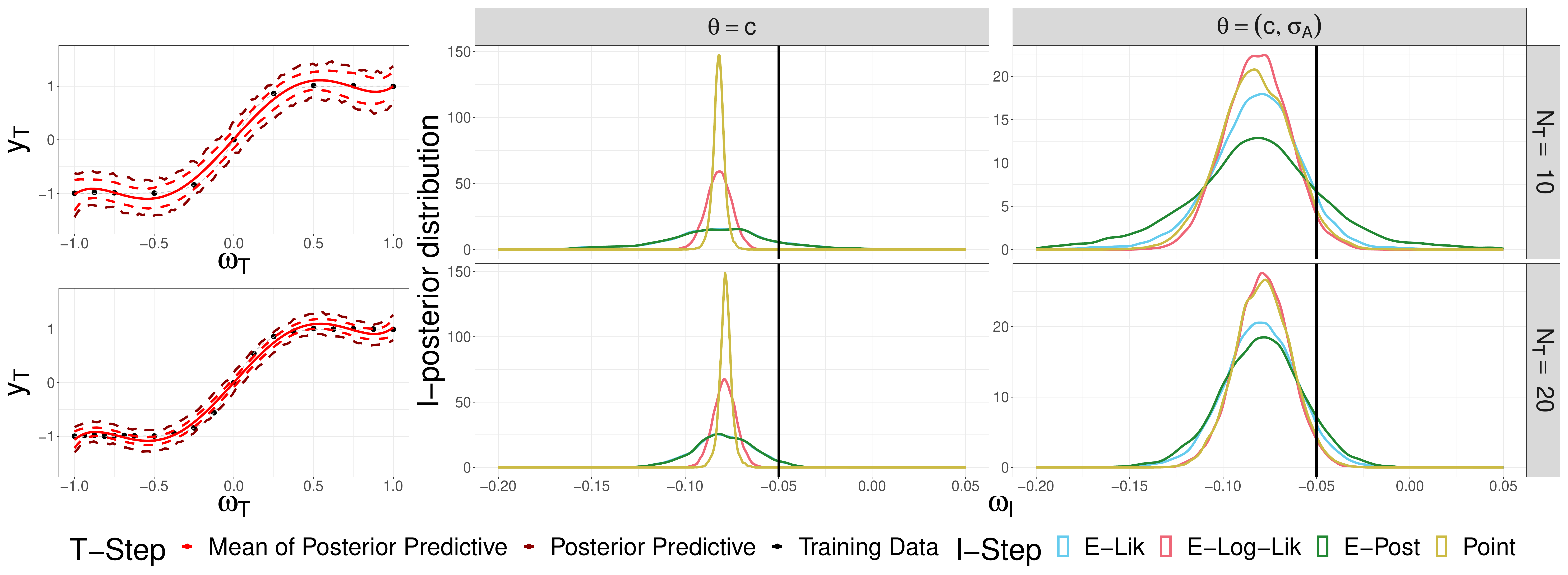}
    \caption{Selected results for two-step procedure with PCE surrogate  in case study 2. Left: For $\Ns = \{10, 20\}$ the training data set $\cD_T$ (black dots), the T-posterior predictive distribution and the mean of the T-posterior predictive distribution (dark red and red lines) is shown. Right: For the true input $\we^* = -0.05$ (black vertical line), we depict the I-posterior distributions for each Point, E-Lik, E-Post, and E-Log-Lik (colored lines). In the center column we only propagate T-epistemic uncertainty via $\theta = c$ and in the right column we propagate both T-epistemic and T-aleatoric uncertainty via $\theta = (c, \sigma_A)$.}    \label{fig:i_step_different_s_steps_pce}
    \end{center}
\end{figure*}

\paragraph{Posterior Distributions}
We perform the T- and I-Step, as previously described in \secref{subsubsection:logistic_true}. During the T-Step, we vary the number of training points $\Ns = \{10, 20\}$ to control for the T-epistemic uncertainty. For the I-Step, we set the true input parameters to $\we^* = -0.05$ and set $\sigma_S = 0.01$ to avoid sampling issues. We show the result of the T-Step in the left column of \figref{fig:i_step_different_s_steps_pce}, where we present both the T-posterior predictive distribution (T-aleatoric and T-epistemic uncertainty) and the mean of the T-posterior predictive distribution (T-epistemic uncertainty only). The T-epistemic uncertainty becomes smaller with increasing number of training data points $N_T$, but the T-aleatoric uncertainty stays approximately constant. 

In the middle column, we show the results of the I-posterior for the four methods when propagating only the T-epistemic uncertainty by considering only the T-posterior draws of the surrogate approximation parameters, i.e. $\theta = c$. For $N_T=10$, the Point I-posterior and the E-Log-Lik I-posterior are narrow despite high T-epistemic uncertainty. In contrast, E-Lik and E-Post produce wider I-posteriors that are similar to each other. As the T-epistemic uncertainty reduces, all methods produce I-posterior distributions that converge towards a similar distribution.

In the right column, we show the I-posteriors when propagating both T-epistemic and T-aleatoric uncertainty by also propagating posterior draws of the surrogate approximation error: $\theta = (c, \sigma_A)$ (see \secref{subsubsection:second_step}). In general, all I-posteriors now tend to be wider than for $\theta = c$ and are more similar to each other. However, in the presence of substantial T-epistemic uncertainty (as for $N_T=10$), E-Post produces the widest I-posterior, followed by E-Lik, Point, and E-Log-Lik. 
In Appendix \ref{subsection:add_pce_post_densities} we provide further results for different true parameter values.
% In S3 we provide further results for different true parameter values.

\begin{figure*}
    \begin{center}
    \includegraphics[width=\textwidth]{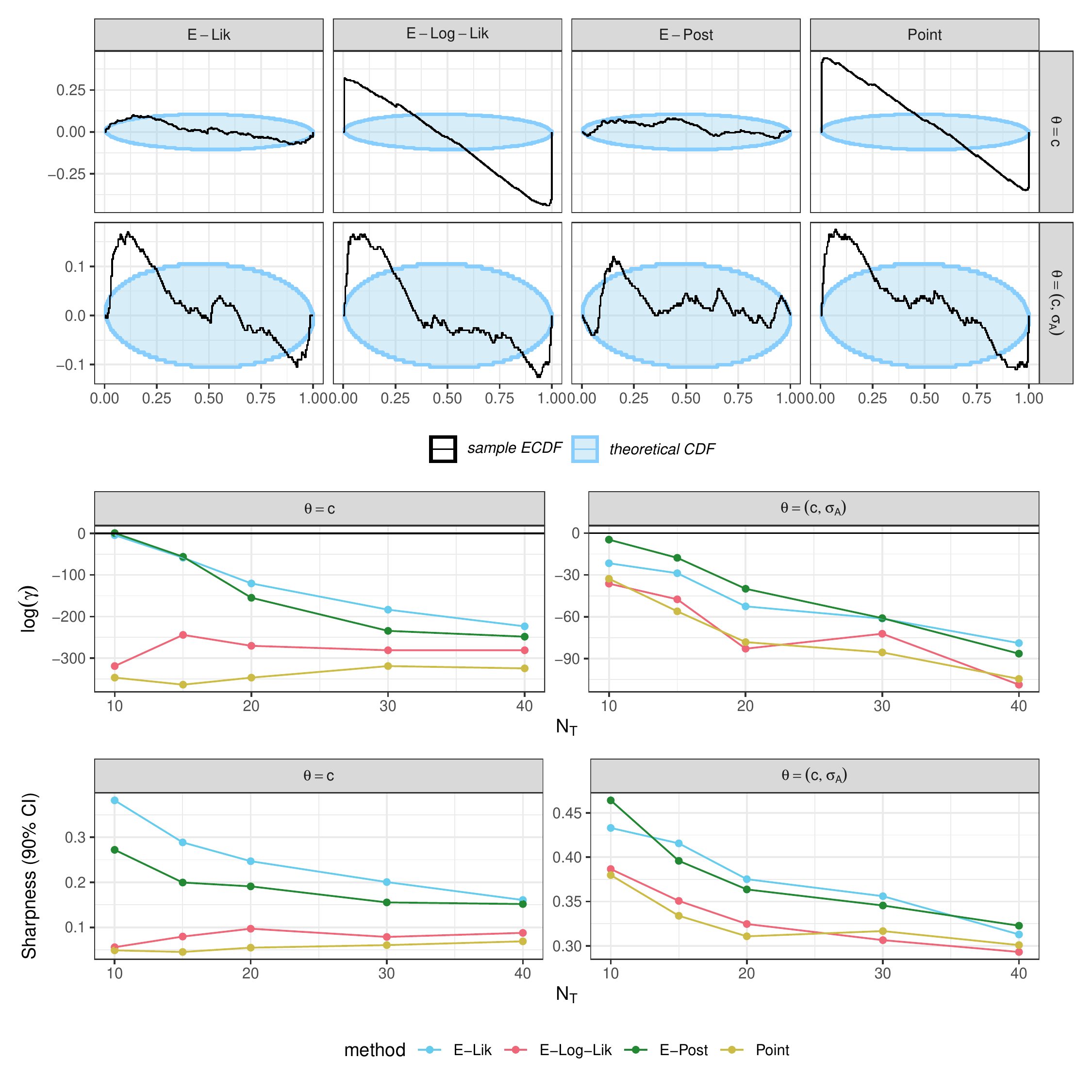}
    \caption{Calibration and sharpness of the I-posteriors using PCE surrogate in case study 2. Top: ECDF difference plots for the I-posterior distributions of $\omega_I$ resulting from the four different methods. We set the number training points $N_T = 10$ and show the results for T-epistemic uncertainty propagation ($\theta = c$) and T-epistemic and T-aleatoric uncertainty propagation ($\theta = (c, \sigma_A)$). Center: log-gamma-statistics of SBC. Bottom: sharpness (90 \% CI) of I-posterior for four different I-Steps (colored dots/lines) for $N_T = \{10, 20, 30, 40, 50\}$.}
    \label{fig:sbc_ecdf_pce}
    \end{center}
\end{figure*}

\paragraph{Calibration}
We performed SBC for our PCE surrogate setup and show the results in \figref{fig:sbc_ecdf_pce}. The top part shows the ECDF-difference plots for $N_T = 10$ for two cases: $\theta=c$ and $\theta = (c, \sigma_A)$. When only T-epistemic uncertainty is propagated, we see that only E-Lik and E-Post produce calibrated results, while E-Log-Lik and Point are overconfident. When T-aleatoric uncertainty is propagated as well, all calibrations improve, while E-Post still produces the best calibration results closely followed by E-Lik. In the bottom of \figref{fig:sbc_ecdf_pce} we compare the calibration via the $\log(\gamma)$-statistic of SBC under varying $N_T = \{10, 20, 30, 40, 50\}$ and confirm the observed pattern described above. As we use a surrogate model which was chosen on purpose to be inflexible and $\sigma_A$ is modelled as constant over $\omega_I$, we cannot achieve proper calibration with either method as epistemic uncertainty reduces.  

\subsection{Case Study 3: Uncertainty Propagation in an SIR Model}\label{subsection:case_study_3}
Finally, we apply our two-step procedure to a real-world case study in epidemiology. The SIR model and its variants are often used to mathematically describe the spread of infectious diseases \citep{hethcote2000mathinfectios, giordano2020modellingcovid}. By considering this model, we demonstrate the applicability of the method to complex real-world problems that lack analytic solutions and require computationally expensive numerical methods. Our approach is particularly useful in such scenarios, as it allows to replace the complex simulation model while propagating relevant surrogate uncertainties.

\paragraph{Setup}
The SIR simulation model $\cM$ is defined through the following system of differential equations:
\begin{align}
\label{eq:SIR}
\begin{split}
    \frac{\mathrm{d}S(t)}{\mathrm{d}t} &= -\beta S(t) \frac{I(t)}{N} \\
    \frac{\mathrm{d}I(t)}{\mathrm{d}t} &= \beta S(t) \frac{I(t)}{N}-\gamma I(t) \\
    \frac{\mathrm{d}R(t)}{\mathrm{d}t} &= \gamma I(t),
\end{split}
\end{align}
where $S(t)$ describes the number of susceptibles, $I(t)$ the number of infectives, and $R(t)$ the number of recovered individuals at time $t$. Furthermore, $\beta$ describes the constant contact rate, $\gamma$ the constant recovery rate, and $N$ denotes the constant population. In the following, we set the constant population to $N = 763$, and fix the initial conditions to the exemplary values $I_0 = 1, S_0 = N - I_0, R_0 = 0$. To solve the differential equation defined in \eqnref{eq:SIR}, we use the Dormand-Prince algorithm \citep{dormand1980family}, a 4th/5th order Runge-Kutta method as implemented in Stan \citep{standev2024stan}.

Typically, measurement data is given for the number of infected individuals, i.e. $y = I(t)$. The unknown parameters to be inferred are $\omega = (\beta, \gamma)$. As a surrogate model, we consider again a PCE (see \secref{subsubsection:logistic_pce}), this time with multivariate Legendre polynomials for the 3-dimensional input $(t, \beta, \gamma)$. The maximum degree of the polynomials is set to $4$, which leads to 34 unknown coefficients $c$ by the standard truncation scheme (see \cite{Sudret2008RESS}). This setup is chosen to demonstrate the applicability of the method to arbitrary complex simulation and surrogate models, as long as samples can be drawn from the posteriors.

For the T-Step, we generate the training data set using a 3-dimensional Sobol sequence \cite{sobol1967} with $N_T = 38$ input points $\{(t^{(i)}, \beta^{(i)}, \gamma^{(i)})\}_{i=1}^{N_T}$. The bounds of the input parameters are $t \in [1, 14]$, $\beta \in [1, 3]$, and $\gamma \in [0.1, 0.9]$. We scale all input parameters linearly to $[-1, 1]$, i.e. the standard scaling of the Legendre polynomials. The output $y_T^{(i)} = I(t^{(i)})$ is then obtained for a given time $t^{(i)}$, $\beta^{(i)}$, and $\gamma^{(i)}$ by solving \eqnref{eq:SIR}. This creates a setup with a low simulation budget, leading to a high T-epistemic uncertainty.

The surrogate parameter priors are chosen in the same way as in \ref{subsubsection:logistic_pce}. To learn the simulation model as efficiently as possible while still enforcing the constraint of a non-negative infection count, we choose a log-normal T-/I-likelihood. The T-model is fitted using NUTS with 4 chains of 1000 warmup and 250 post-warmup sampling iterations, resulting in a total of $S=1000$ samples to propagate.

For the I-Step, we generate $N_I = 50$ measurements by sampling output responses $y_T^{(i)} \sim \text{NegBin}(I(t), \phi)$ given evenly spaced $t$, true input parameters $\omega_I^*$, and the shape parameter $\phi = 9.6$. As priors we set $\beta \sim \cN_{[1, 3]}(2, 0.5^2)$ and $\gamma \sim \cN_{[0.1, 0.9]}(0.5, 0.25^2)$ in order to stay within the parameter bounds that were used to train the surrogate model and further set $\sigma_I \sim \text{Half-Normal}( 0.5^2)$. The densities and parameterizations of the used distributions are provided in \secref{subsection:negbinom_lonormal}.

In this setup, we propagate epistemic uncertainty contained in $c$ through the $S = 1000$ T-posterior samples. For the Point, E-Lik, and E-Log-Lik methods, we sample from the I-posterior of $\we$ and $\se$ using NUTS with 32 chains, each running for 1000 warmup and 1000 post-warmup iterations, resulting in a total of 32000 I-posterior draws. Instead, for E-Post, where we fit $S=1000$ separate models (one for each T-posterior draw), we use NUTS with 4 chains, each running for 250 iterations, resulting in a total of $10^6$ posterior draws.

\begin{figure*}
    \begin{center}
    \includegraphics[width=\textwidth]{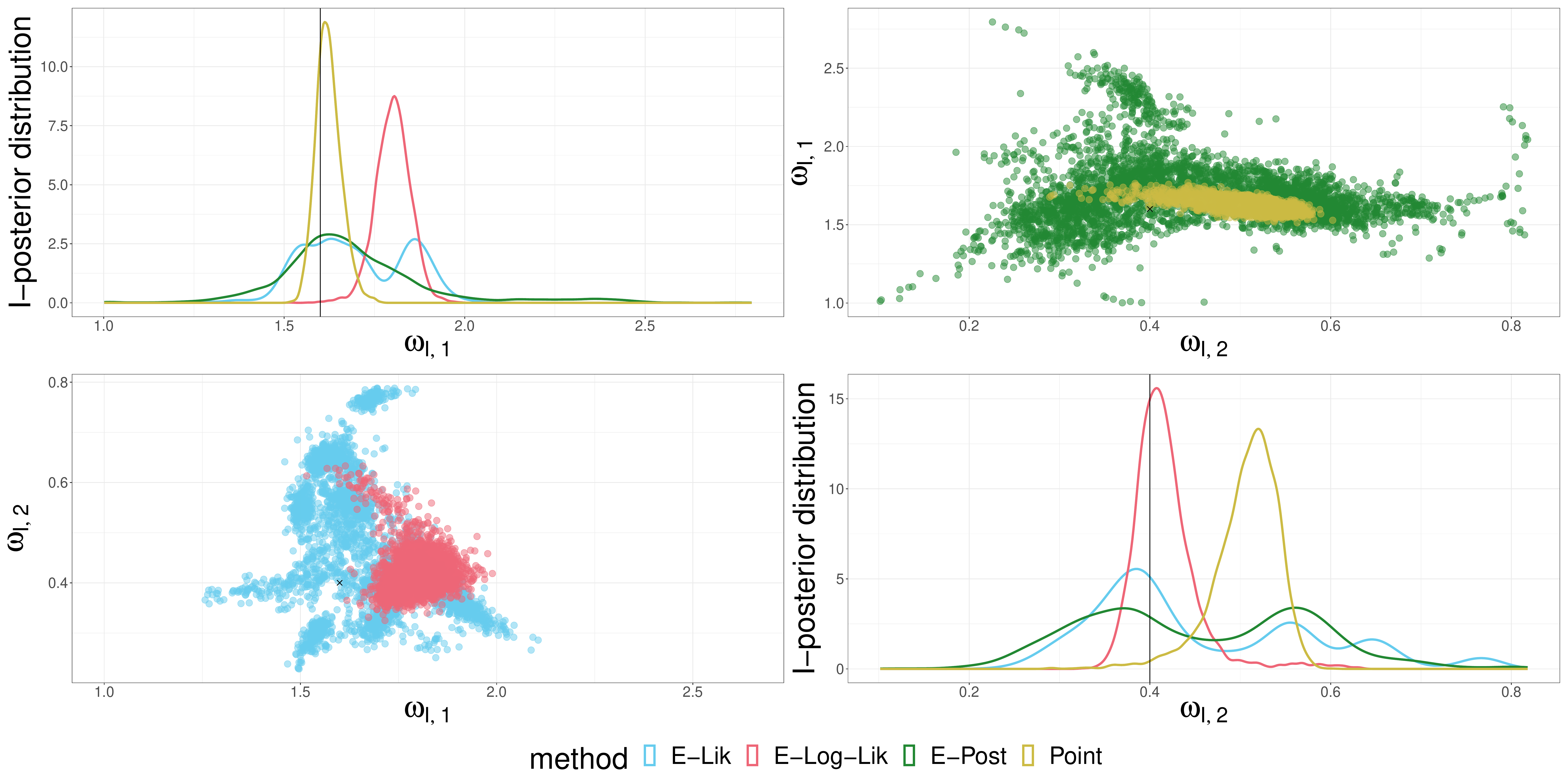}
    \caption{Pairs plot of the I-posteriors using a PCE surrogate in case study 3. On the diagonals we depict for the true input $\omega_I^* = (\beta^*, \gamma^*) = (1.6, 0.4)$ (black vertical line) the marginal I-posterior distributions for the Point, E-Lik, E-Post, and E-Log-Lik methods (colored lines). The off-diagonals show the scatter plots of 5000 randomly subsampled I-posterior draws. We propagate T-epistemic uncertainty via $\theta = c$.}
    \label{fig:case_study_3_i_posteriors}
    \end{center}
\end{figure*}

\paragraph{Results}
We set the underlying true input parameters to exemplary values $\omega_I^* = (\omega^*_{I, 1}, \omega^*_{I, 2}) = (\beta^*, \gamma^*) = (1.6, 0.4)$ and compare the I-posteriors resulting from the four different methods in \figref{fig:case_study_3_i_posteriors}. In the marginal I-posterior distribution plots, we observe that both the Point and E-Log-Lik methods produce an I-posterior that is overconfident in one dimension. We also observe that both the E-Lik and E-Post methods have substantial probability mass around the true values. In the scatter plots showing the joint I-posterior distribution, we observe that Point and E-Log-Lik produce samples that do not cover the true value $\omega_I^*$, while E-Post and E-Lik do. The large uncertainty produced by these methods (E-Post and E-Lik) resembles the uncertainty in the T-posterior, as the surrogate was only trained on very limited data.

\section{Conclusion}
We introduced a general two-step procedure for surrogate-based Bayesian inference. Within our approach, we propagate all relevant surrogate uncertainties (both aleatoric and epistemic) from the surrogate training to the real-data inference step, thereby producing fully uncertainty aware inference on the real data. The uncertainty propagation methods developed within our framework are in principle agnostic to the chosen surrogate, but require the ability to sample from its training-step posterior given data generated from the simulator. 
To evaluate the uncertainty calibration of the resulting inference, we proposed an extension of simulation-based calibration suitable for our two-step surrogate approach.

As we demonstrate in our case studies, even in seemingly simple setups, complex behavior occurs in terms of posterior shape and calibration when propagating surrogate uncertainty. In particular two uncertainty propagation methods (E-Lik \& E-Post) showed substantial improvements in uncertainty calibration compared to traditional surrogate-based inference that is uncertainty-unaware. 
What is more, our results show the importance of propagating the complete surrogate uncertainty (aleatoric and epistemic) instead of propagating only parts of it.
Intuitively, one might expect that there is only one ``correct'' uncertainty propagation method within the bounds of probability theory. However, as we demonstrate, the two advocated methods for uncertainty propagation (E-Lik \& E-Post) produce non-equivalent inference even in simple cases despite being equally justified by probability theory. That said, at least in our case studies, the produced inference was very similar. They differ in the computational requirements and ease of parallelization though (see Section \ref{parallelization}), such that either or the other may be preferable depending the context and available resources.

\paragraph{Future Work}
Our surrogate-based Bayesian inference approach is agnostic to the input-parameter dimensionality and its scaling is unaffected by said dimensionality (but only by the number of draws propagated from the training to the inference step). In our case studies, we focused on simulators with up to three-dimensional input parameters in order to simplify the presentation and establish a better intuition about the overall approach. To study the applicability of our methods on high-dimensional challenges, biological systems requiring accurate, uncertainty-aware inference \citep{mitra2019peuqssysbio} will offer an interesting class of problems for future research. 

In our case studies, we modelled the surrogate approximation error to be constant, thus assuming the aleatoric uncertainty of the surrogate to be independent of the input parameters. However, this assumption is likely unjustified in practice, if there is substantial approximation error due to the surrogate inflexibility. Hence, one could improve the modeling of the surrogate approximation error by conditioning it on the input parameters, for example, as suggested in \cite{kohlhaas2023pceactive}. This will induce additional parameters to model the approximation error, which are then seamlessly propagated through our two-step procedure. 

Our current implementation of uncertainty quantification relies on MCMC in both training and inference step. This may become computationally expensive, sometimes prohibitively so, as the number of (surrogate) model parameters grows \citep{Izmailov2021whatarebayesiannnlike}. Hence, for a surrogate model with hundreds of thousands of parameters, full Bayesian UQ will become intractable and approximations are needed. Approaches like variational inference \citep{hinton1993keepingthennsimple, graves2011practicalvinn} and partial UQ on the last neural network layer \citep{kristiadi2020bayesoverconfidencerelu, FiedlerL2023improveduqfornnwbll, harrison2024vbll} could be sensible alternatives. Our uncertainty propagation methods only require the ability to draw samples from an (approximate) posterior in the training step that can be subsequently passed to the inference step. This flexibility ensures that our approach is still applicable in scenarios where a full Bayesian UQ is infeasible, but the implication on inference validity, in particular uncertainty calibration, need to be further studied.

\section*{Acknowledgments}
Partially funded by Deutsche Forschungsgemeinschaft (DFG, German Research Foundation) under Germany's Excellence Strategy EXC 2075 – 390740016 and DFG Project 500663361. We acknowledge the support by the Stuttgart Center for Simulation Science.

\bibliographystyle{apalike}

\bibliography{bibliography}

\appendix
\section{Notation}
\begin{itemize}
    \item $\omega$: input
    \item $y$: output
    \item $\cM$: simulator
    \item $\widetilde{\cM}$: surrogate model
    \item $c$: surrogate approximation parameters
    \item $d$: number of surrogate parameters
    \item T-Step: surrogate training step
    \begin{itemize}
        \item $\Ns$: number of simulation training pairs
        \item $\ws$: simulation input
        \item $\ys$: simulation output
        \item $e_S$: simulation noise
        \item $\sigma_S$: simulation noise hyperparamters
        \item $e_A$: approximation error
        \item $\sigma_A$: surrogate approximation error hyperparameters
        \item $\theta = (c, \sigma_A)$: trainable surrogate parameters
        \item $S$: number of T-posterior samples 
    \end{itemize}
    \item I-Step: surrogate inference step
    \begin{itemize}
        \item $\Ne$: number of measurements
        \item $\we$: quantity of interest (QoI)
        \item $\ye$: measurement data
        \item $e_I$: measurement noise
        \item $\sigma_I$: measurement noise hyperparameters
        \item $K$: number of I-posterior samples
    \end{itemize}
\end{itemize}
\subsection{Negative binomial, Log-normal, and Half-normal distributions}
\label{subsection:negbinom_lonormal}
Below, we show the Negative binomial, Log-normal and Half-normal distributions. The probability mass function of the Negative binomial distribution for scalar count $n \in \nN$ with the two positive parameters $\mu \in \nR^+$ and $\phi \in \nR^+$ is given by:
\begin{align}
\begin{split}
    &p_{\text{NegBinom}}(n \mid \mu, \phi) = \\
    &\binom{n + \phi - 1}{n} \left( \frac{\mu}{\mu + \phi} \right)^n \left( \frac{\phi}{\mu + \phi} \right)^\phi.
\end{split}
\end{align}
In this parameterization the mean and variance are given by
\begin{align}
    \nE[n] = \mu \quad \text{and} \quad \text{Var}[n] = \mu + \frac{\mu^2}{\phi}.
\end{align}
The probability density function of the Log-normal distribution for a positive scalar $y \in \nR^+$ with the parameters $\mu \in \nR$ and $\sigma \in \nR^+$ is given by:
\begin{align}
\begin{split}
    &p_{\text{Log-Normal}}(y \mid \mu, \sigma) \\
    &=\frac{1}{\sqrt{2 \pi} \sigma} \frac{1}{y} \exp \left( - \frac{1}{2} \left( \frac{\log y - \mu }{\sigma} \right)^2 \right).
\end{split}
\end{align}
In this parameterization the mean and variance are given by
\begin{align}
    &\nE[y] = \exp \left(\mu + \frac{\sigma^2}{2} \right) \quad \text{and} \\
    &\text{Var}[y] = [\exp(\sigma^2) - 1] \exp(2 \mu + \sigma^2).
\end{align}
The probability density function of the Half-normal distribution for a positive scalar $y \in \nR^+$ with the parameter $\sigma \in \nR^+$ is given by:
\begin{align}
\begin{split}
    &p_{\text{Half-Normal}}(y \mid \sigma^2) =\frac{\sqrt{2}}{\sqrt{\pi} \sigma} \exp \left( - \frac{y^2}{2\sigma^2} \right).
\end{split}
\end{align}
\section{Derivations}

\subsection{Inequality of E-Lik and E-Post via counterexample}\label{subsection:counterexample}
% In the following, we show the inequality of E-Post (\secref{subsubsection:E-Post}) and E-Lik (\secref{subsubsection:E-Lik}) using a counterexample with discrete random variables. 
In the following, we show the inequality of E-Post and E-Lik (Main Section 2.2) using a counterexample with discrete random variables. 
We denote the T-posterior as $p(\theta)$ and omit the dependence on the training data $\cD_T$ to simplify notation.

Let $\omega \in \{0, 1\}$, $y \in \{0, 1\}$, $\theta \in \{0, 1\}$. We set the probability values
\[
\begin{array}{c|c}
 &  \\
\hline
p(\omega = 0) & 1/2 \\
p(\omega = 1) & 1/2 
\end{array}
\qquad
\begin{array}{c|c}
 &  \\
\hline
p(\theta = 0) & 1/2 \\
p(\theta = 1) & 1/2 
\end{array}
\]
\[
\begin{array}{c|c|c}
 & \omega=0, \theta=0 & \omega=1, \theta=0 \\
\hline
p(y = 0 \mid \omega, \theta) & 1/4 & 1/2 \\
p(y = 1 \mid \omega, \theta) & 3/4  & 1/2
\end{array}
\]
\[
\begin{array}{c|c|c}
 & \omega=0, \theta=1 & \omega=1, \theta=1 \\
\hline
p(y = 0 \mid \omega, \theta) & 1/2 & 1/2 \\
p(y = 1 \mid \omega, \theta) & 1/2 & 1/2
\end{array}
\]
First, we state the general formulation of E-Lik and E-Post for discrete pmf's:
\begin{align*}
    &p(\omega \mid y, u=\text{E-Lik}) \\
    &= \frac{p(\omega)\sum_i p(y \mid \omega, \theta=i)p(\theta=i)}{\sum_i \sum_j p(y \mid \omega=j, \theta=i)p(\omega=j)p(\theta=i)} \\
    &p(\omega \mid  y, u=\text{E-Post}) \\
    &= \sum_i\frac{p(\omega) p(y \mid \omega, \theta=i)p(\theta=i)}{\sum_j p(y \mid \omega=j, \theta=i)p(\omega=j)}\\
    &\times\frac{p(\omega=0) p(y=0 \mid \omega=0, \theta=1)p(\theta=1)}{\sum_j p(y \mid \omega=j, \theta=1)p(\omega=j)}
\end{align*}
We calculate E-Post for $\omega=0$ and $y=0$ by plugging in the probability values:
\begin{align*}
    &p(\omega=0 \mid  y=0, u=\text{E-Post}) \\
    &= \sum_i\frac{p(\omega=0) p(y=0 \mid \omega=0, \theta=i)p(\theta=i)}{\sum_j p(y=0 \mid \omega=j, \theta=i)p(\omega=j)}\\
    &= \frac{1/2 \cdot 1/4 \cdot 1/2}{3/8} + \frac{1/2 \cdot 1/2 \cdot 1/2}{1/2} = \frac{1/16}{3/8} + \frac{1/8}{1/2} \\
    &= \frac{5}{12} \approx 0.417
\end{align*}
Next, we calculate E-Lik for $\omega=0$ and $y=0$:
\begin{align*}
    &p(\omega=0 \mid  y=0, u=\text{E-Lik}) \\
    &= \frac{p(\omega=0)\sum_i p(y=0 \mid \omega=0, \theta=i)p(\theta=i)}{\sum_i \sum_j p(y=0 \mid \omega=j, \theta=i)p(\omega=j)p(\theta=i)}\\
    &= \frac{1/2 \cdot (1/4 \cdot 1/2 + 1/2 \cdot 1/2)}{7/16} \\
    &= \frac{3}{7}  \approx 0.429.
\end{align*}
We see that the results produced by E-Post and E-Lik for $\omega=0$ and $y=0$ are similar, but not equal.

The normalization constants for E-Post were given by: 
\begin{align*} 
    &\sum_j p(y \mid \omega=j, \theta=0)p(\omega=j) \\
    &= p(y=0 \mid \omega=0, \theta=0)p(\omega=0)\\
    &+p(y=0 \mid \omega=1, \theta=0)p(\omega=1) \\
    &= 1/4 \cdot 1/2+ 1/2 \cdot 1/2 = 3/8
\end{align*}
and 
\begin{align*}
    &\sum_j p(y \mid \omega=j, \theta=1)p(\omega=j) \\
    &= p(y=0 \mid \omega=0, \theta=1)p(\omega=0)\\
    &+p(y=0 \mid \omega=1, \theta=1)p(\omega=1) \\
    &= 1/2 \cdot 1/2+ 1/2 \cdot 1/2 = 1/2.
\end{align*}
The normalization constant for E-Lik was given by:
\begin{align*}
    &p_{\text{norm}}(y=0) \\
    =& \sum_i \sum_j p(y=0 \mid \omega=j, \theta=i)p(\omega=j)p(\theta=i) \\
    =&  p(y=0 \mid \omega=0, \theta=0)p(\omega=0)p(\theta=0) \\
    &+ p(y=0 \mid \omega=1, \theta=0)p(\omega=1)p(\theta=0) \\
     &+ p(y=0 \mid \omega=0, \theta=1)p(\omega=0)p(\theta=1) \\
     &+ p(y=0 \mid \omega=1, \theta=1)p(\omega=1)p(\theta=1)\\
    =& 1/4 \cdot 1/2 \cdot 1/2 + 1/2 \cdot 1/2 \cdot 1/2 \\
    &+ 1/2 \cdot 1/2 \cdot 1/2 + 1/2 \cdot 1/2 \cdot 1/2 = 7/16
\end{align*}

\subsection{Equivalence of E-Log-Lik and E-Log-Post}\label{subsection:e-log-post}
% Here we show that the two formulations E-Log-Post and E-Log-Lik (\secref{subsubsection:E-Log-Lik}) are equivalent. 
Here we show that the two formulations E-Log-Post and E-Log-Lik (Main Section 2.2) are equivalent. 
The E-Log-Lik is defined as:
\begin{align*}
    &\log(p(\we \mid \ye, u=\text{E-Log-Lik})) \\
    &\propto \log(p(\we)) + \int \log(p(\ye \mid \we, \theta))p(\theta \mid \cD_T) d\theta
\end{align*}
Similarly, to E-Post and E-Lik we can define the E-Log-Post:
\begin{align*}    &\log(p(\we \mid \ye, u=\text{E-Log-Post})) \\
    &= \int \log(p(\we \mid \ye)) p(\theta \mid \ys)d\theta \\
    &=\int \left[ \log(p(\we)) + \log(p(\ye \mid \we, \theta) - \log(C(\theta)) \right]\\
    &\times p(\theta \mid \ys) d\theta \\
    &= \log(p(\we)) \int p(\theta \mid \ys) d\theta \\
    &+ \int \log(p(\ye \mid \we, \theta)) p(\theta \mid \ys) d\theta \\
    &- \int \log(C(\theta)) p(\theta \mid \ys) d\theta \\
    &\propto \log(p(\we)) + \int \log(p(\ye \mid \we, \theta)) p(\theta \mid \ys) d\theta
\end{align*}
In the last equation we used that the integral over a probability distribution is one, i.e. $\int p(\theta \mid \ys)d\theta = 1$ and since the posterior is a function of $\we$ we can define a new constant $C_1(\theta) := - \int \log(C(\theta)) p(\theta \mid \ys) d\theta$.

\subsection{Case Study 1: Slope Intercept Model}
% In this section we derive the E-Log-Lik for the slope intercept model stated in \secref{subsection:linear_model}. 
In this section we derive the E-Log-Lik for the slope intercept model stated in Main Section 3.1.
% Further we show the E-Lik with full normalization constant.
Let $c = [c_1, c_2]^T$, 
$$\Omega_T = \begin{bmatrix}
1 & \omega_T^{(1)} \\
1 & \omega_T^{(2)} \\
\vdots & \vdots \\
1 & \omega_T^{(N_T)} 
\end{bmatrix},$$
$\hat{\omega}_I = [1, \omega_I]^T$, 
$\mu_{T1} = [\mu_{T1}^{(1)}, \mu_{T2}^{(2)}]^T$,
$$\Sigma_{T1} = \begin{bmatrix}
\Sigma_{T1}^{(1, 1)} & \Sigma_{T1}^{(1, 2)}\\
\Sigma_{T1}^{(2, 1)} & \Sigma_{T1}^{(2, 2)}
\end{bmatrix},$$

\subsubsection{E-Log-Lik}\label{subsubsection:derivation_e-log-lik}
Now we derive the stated Expected-Log-Likelihood result:
\begin{align*}
    &p(\omega_I \mid y_I, u=\text{E-Log-Lik})\\
    \propto {}&p(\omega_I)\exp\left\{\int \log(p(y_I \mid \omega_I, c))p(c \mid \cD_T) dc\right\} \\
    ={}& p(\omega_I) \exp\left\{\int \log(\mathcal{N}(y_I \mid \hat{\omega}_I^\top c, \sigma_{I}^2))\right.\\
    &\left.\times\mathcal{N}(c \mid \mu_{T1}, \Sigma_{T1})dc\right\}\\
    ={}& p(\omega_I) \exp\left\{\int -\frac{1}{2\sigma_{I}^2}(\hat{\omega}_I^\top c-y_I)^2\right.\\
    &\left.\times\mathcal{N}(c \mid \mu_{T1}, \Sigma_{T1})dc\right\} \\
    ={}& p(\omega_I) \exp\left\{-\frac{1}{2\sigma_{I}^2} \int (y_I^2 - 2y_I \hat{\omega}_I^\top c+c^\top \hat{\omega}_I \hat{\omega}_I^\top c)\right.\\
    &\left.\times\mathcal{N}(c \mid \mu_{T1}, \Sigma_{T1})dc\right\} \\
    ={}& p(\omega_I) \exp\left\{-\frac{1}{2\sigma_{I}^2} (y_I^2 - 2y_I \hat{\omega}_I^\top \nE[c]\right.\\
    &\left.+\nE[c^\top \hat{\omega}_I \hat{\omega}_I^\top c])\right\}\\
    ={}& p(\omega_I) \exp\left\{-\frac{1}{2\sigma_{I}^2} (y_I^2 - 2y_I \hat{\omega}_I^\top \mu_{T1}\right.\\
    &\left.+\text{Tr}(\hat{\omega}_I\hat{\omega}_I^\top \Sigma_{T1})+\mu_{T1}^\top\hat{\omega}_I \hat{\omega}_I^\top \mu_{T1})\right\} \\
    \begin{split}
        \propto{}& p(\omega_I) \exp\left\{-\frac{1}{2\sigma_{I}^2} (y_I^2 - 2y_I \hat{\omega}_I^\top \mu_{T1}+\omega_I^2\Sigma_{T1}^{(2,2)}\right.\\
        &+\omega_I(\Sigma_{T1}^{(2,1)}+\Sigma_{T1}^{(1,2)}) + \mu_{T1}^{(2)}\mu_{T1}^{(2)}\omega_I^2\\
        &\left.+2\mu_{T1}^{(1)}\mu_{T1}^{(2)}\omega_I)\right\}
    \end{split}\\
    \propto{}& p(\omega_I) \exp \left\{-\frac{1}{2\sigma_{I}^2} (\mu_{T1}^{(2)}\mu_{T1}^{(2)}+\Sigma_{T1}^{(2,2)})\right.\\
    &\left.\times(\omega_I-\frac{y\mu_{T1}^{(2)}-\Sigma_{T1}^{(1,2)}-\mu_{T1}^{(1)}\mu_{T1}^{(2)}}{\mu_{T1}^{(2)}\mu_{T1}^{(2)}+\Sigma_{T1}^{(2,2)}})^2\right\} \\
    \propto{}& \cN(\omega_I \mid  \mu_{I1}, \sigma_{I1}^2),
\end{align*}
with
\begin{align*}
    \sigma_{I1}^2 ={}& (\sigma_{I0}^{-2} +\se^{-2} (\mu_{T1}^{(2)}\mu_{T1}^{(2)} + \Sigma_{T1}^{(2,2)}))^{-1}, \\
    \mu_{I1} ={}& \sigma_{I1}^2(\sigma_{I0}^{-2}\mu_{I0} + \se^{-2}(\mu_{T1}^{(2)}\ye-\Sigma_{T1}^{(1,2)}-\mu_{T1}^{(1)}\mu_{T1}^{(2)}))
\end{align*}
where we used \citep[p.~43]{petersen2012matrixcookbook}:
\begin{align*}
    &\text{Tr}(\hat{\omega}_I \hat{\omega}_I^\top \Sigma_{T1}) \\
    &= \omega_I^2\Sigma_{T1}^{(2,2)}+\omega_I(\Sigma_{T1}^{(2,1)}+\Sigma_{T1}^{(1,2)})+\Sigma_{T1}^{(1,1)}
\end{align*}
and
\begin{align*}
    &\mu_{T1}^\top \hat{\omega}_I \hat{\omega}_I^\top \mu_{T1}\\
    &= \mu_{T1}^{(2)}\mu_{T1}^{(2)}\omega_I^2+2\mu_{T1}^{(1)}\mu_{T1}^{(2)}\omega_I+\mu_{T1}^{(1)}\mu_{T1}^{(1)}
\end{align*}
%\subsubsection{E-Lik}\label{subsubsection:derivation_e-lik}
%E-Lik with normalization constant reads as:
%\begin{align*}
%    &p(\we \mid \ye, u=\text{E-Lik}) \\
%    &= \frac{p(\we) \int p(\ye \mid \we, c) p(c \mid \cD_T) dc}{\int p(\we) \int p(\ye \mid \we, c) p(c \mid \cD_T) dc d\we}\\
%    &= \frac{\mathcal{N}(\omega_I \mid \mu_{I0}, \sigma_{I0}^2)\int \mathcal{N}(y_I \mid  \hat{\omega}_I^\top c, \sigma_I^2)\mathcal{N}(c \mid \mu_{T1}, \Sigma_{T1})dc}{\int\mathcal{N}(\omega_I \mid \mu_{I0}, \sigma_{I0}^2)\int \mathcal{N}(y_I \mid  \hat{\omega}_I^\top c, \sigma_I^2)\mathcal{N}(c \mid \mu_{T1}, \Sigma_{T1})dcd\omega_I} \\
%    &= \frac{\mathcal{N}(\omega_I \mid \mu_{I0}, \sigma_{I0}^2)\mathcal{N}(y_I \mid \hat{\omega}_I^\top  \mu_{T1}, \hat{\omega}_I^\top\Sigma_{T1}\hat{\omega}_I+\sigma_{I}^2)}{\int \mathcal{N}(\omega_I \mid \mu_{I0}, \sigma_{I0}^2)\mathcal{N}(y_I \mid \hat{\omega}_I^\top  \mu_{T1}, \hat{\omega}_I^\top\Sigma_{T1}\hat{\omega}_I+\sigma_{I}^2) d\we }
%\end{align*}

\section{Further results}
\subsection{Case Study 1: Slope only Model}
%In \secref{subsection:linear_model} we showed the I-posteriors for the linear model with a slope and an intercept. 
In Main Section 3.1 we showed the I-posteriors for the linear model with a slope and an intercept. 
In this section, we consider an even simpler setup, where the simulator and surrogate model are both a slope only model:
\begin{align}
    y = \cM(\omega) = \widetilde{\cM}(\omega, c) = c\cdot \omega,
\end{align}
where $\omega \in \nR$ is the input, $y \in \nR$ the output and $c \in \nR$ is the unknown parameter.
We consider to propagate only the surrogate approximation parameter, i.e. we set $\theta=c$. We set normal priors and normal likelihoods.
\paragraph{T-Step}
As our simulation data set we use one input-output pair $\cD_T = \{\omega_T, y_T\}$ and $N_T = 1$.
We set a normal prior on $c$ with mean $\mu_{T0}$ and variance $\sigma_{T0}^2$. We calculate the posterior using the conjugate prior relation for a normal-normal model \citep{murphy2007conjugate, pml1Book}:
\begin{align}
p(c \mid y_T) &= \frac{p(c)p(y_T \mid c)}{p(y_T)}\\
\begin{split}
             &=\mathcal{N}(c \mid \mu_{T1}, \sigma_{T1}^2), \\
\sigma_{T1}^2 &= (\sigma_{T0}^{-2} +\sigma_A^{-2}\omega_T^2)^{-1}, \\
\mu_{T1} &= \sigma_{T1}^2(\sigma_{T0}^{-2}\mu_{T0} + \sigma_A^{-2}\omega_T y_T)
\end{split}
\end{align}
\paragraph{I-Step: Point}
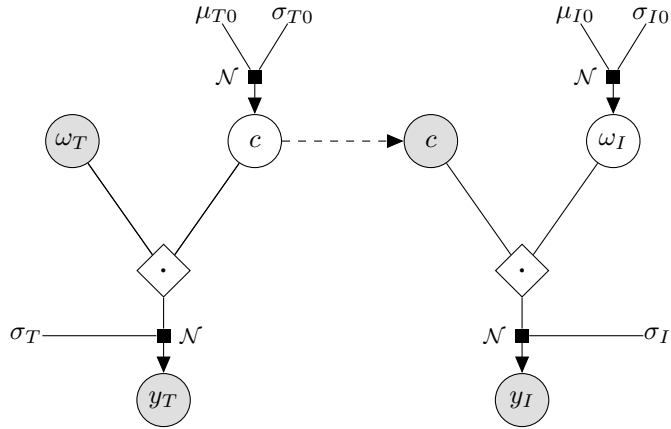
\begin{figure*}[ht]
  \begin{center}
    \begin{tabular}{cc}
      \begin{tikzpicture}

  \node[obs]                               (y_S) {$\ys$};
  \factor[above=of y_S] {y_S-f} {right:$\mathcal{N}$} {} {} ; 
  \node[det, above=of y_S]            (dot_S) {$\boldsymbol{\cdot}$} ; 
  \node[obs, above=of dot_S, xshift=-1.2cm]    (w_S) {$\ws$};
  \node[latent, above=of dot_S, xshift=1.2cm]  (c) {$c$};
  \node[obs, right=4cm of y_S]               (y_I) {$\ye$};
  \node[det, above=of y_I]            (dot_I) {$\boldsymbol{\cdot}$} ; 
  \factor[above=of y_I] {y_I-f} {left:$\mathcal{N}$} {} {} ; 
  \node[latent, above=of dot_I, xshift=1.2cm]    (w_I) {$\we$};
  \node[obs, above=of dot_I, xshift=-1.2cm]  (c_S) {$c$};

  \node[const, above=1.2 of c, xshift=-0.5cm] (ms) {$\mu_{T0}$} ; 
  \node[const, above=1.2 of c, xshift=0.5cm]  (ss) {$\sigma_{T0}$} ; 

  \node[const, above=1.2 of w_I, xshift=-0.5cm] (mi) {$\mu_{I0}$} ; %
  \node[const, above=1.2 of w_I, xshift=0.5cm]  (si) {$\sigma_{I0}$} ; 

  \node[const, left=1.5cm of y_S-f] (ssl) {$\sigma_{T}$} ; 
  \node[const, right=1.5cm of y_I-f] (sil) {$\sigma_{I}$} ; 
  
  \factor[above=of c] {c-f} {left:$\mathcal{N}$} {ms,ss} {c} ; 
  \factor[above=of w_I] {w_I-f} {left:$\mathcal{N}$} {mi,si} {w_I} ; 
  \factoredge {dot_S,ssl} {y_S-f} {y_S} ; 
  \factoredge {dot_I,sil} {y_I-f} {y_I} ; 
  
  \edge[-] {w_S,c} {dot_S} ;
  \edge[dashed] {c} {c_S} ;

  \edge[-] {w_S,c} {dot_S} ;
  \edge[-] {w_I,c_S} {dot_I} ;

\end{tikzpicture}
    \end{tabular}
  \end{center}
  \caption{Two-Step-Procedure for linear model with normal likelihoods and normal priors.}
  \label{fig:linear_graphical_model}
\end{figure*}
We set a normal prior on $\omega_I$:
\begin{align}
    p(\omega_I) = \mathcal{N}(\omega_I \mid \mu_{I0}, \sigma_{I0}^2).
\end{align}
We use the following normal likelihood:
\begin{align}
    p(y_I \mid \omega_I, c) = \mathcal{N}(y_I \mid c \cdot \omega_I, \sigma_{I}^2)
\end{align}
We compute the mean of the T-Step posterior: $\bar{c} = \mu_{T1}$
Now we can compute the Point I-posterior using the conjugate prior relation \citep{pml1Book}:
\begin{align*}
    &p(\we \mid \ye, u=\text{Point}) \\
    &\propto  p(\omega_I)p(y_I \mid \omega_I, \bar{c})\\
    &= \mathcal{N}(\omega_I \mid \mu_{I0}, \sigma_{I0}^2)  \mathcal{N}(y_I \mid \mu_{T1} \cdot \omega_I, \sigma_{I}^2) \\
    \begin{split}
        &= \cN(\we \mid  \mu_{I1}, \sigma_{I1}^2) \\
        &\text{, with:}\\
        \sigma_{I1}^2 &= (\sigma_{I0}^{-2} +\se^{-2} \mu_{T1}^2)^{-1}, \\
        \mu_{I1} &= \sigma_{I1}^2(\sigma_{I0}^{-2}\mu_{I0} + \se^{-2}\mu_{T1}\ye)
    \end{split}
\end{align*}
\paragraph{I-Step: E-Lik}
Derivation
\begin{align*}
    &p(\we \mid \ye, u=\text{E-Lik})\\
    &\propto p(\we) \int p(\ye \mid \we, c) p(c \mid \cD_T) dc\\
    &= \mathcal{N}(\omega_I \mid \mu_{I0}, \sigma_{I0}^2)\\
    &\times \int \mathcal{N}(y_I \mid \omega_I \cdot c, \sigma_I^2)\mathcal{N}(c \mid \mu_{T1}, \sigma_{T1}^2)dc \\
    &= \mathcal{N}(\omega_I \mid \mu_{I0}, \sigma_{I0}^2)\mathcal{N}(y_I \mid \omega_I \cdot \mu_{T1}, \omega_I^2\cdot\sigma_{T1}^2+\sigma_{I}^2)
\end{align*}
\paragraph{I-Step: E-Log-Lik}
\begin{align*}
    &p(\we \mid \ye, u=\text{E-Log-Lik})\\
    &\propto p(\we)\exp\left\{\int \log(p(\ye \mid \we, c))p(c \mid \cD_T) dc\right\} \\
    &= \mathcal{N}(\omega_I \mid \mu_{I0}, \sigma_{I0}^2)\exp\left\{\int \log(\mathcal{N}(y_I \mid \omega_I \cdot c, \sigma_{I}^2))\right.\\
    &\left.\times\mathcal{N}(c \mid \mu_{T1}, \sigma_{T1}^2)dc\right\} \\
    &\propto \mathcal{N}(\omega_I \mid \mu_{I0}, \sigma_{I0}^2)\exp\left\{\int \left(\frac{1}{2\sigma_I^2} (y_I-\omega_I c)^2\right)\right.\\
    &\left.\times\mathcal{N}(c \mid \mu_{T1}, \sigma_{T1}^2) dc \right\} \\
    \begin{split}
        &= \cN(\we \mid \mu_{I1}, \sigma_{I1}^2) \\
        &\text{, with:}\\
        \sigma_{I1}^2 &= (\sigma_{I0}^{-2} +\se^{-2} (\mu_{T1}^2+\sigma_{T1}^2))^{-1}, \\
        \mu_{I1} &= \sigma_{I1}^2(\sigma_{I0}^{-2}\mu_{I0} + \se^{-2}\mu_{T1}\ye)
    \end{split}
\end{align*}
\paragraph{I-Step: E-Post}
\begin{align*}
    &p(\we \mid \ye, u=\text{E-Post}) \\
    &= p(\omega_I) \int \frac{p(y_I \mid \omega_I, c)p(c \mid \cD_T)}{\int p(y_I \mid  \omega_I, c) p(\omega_I) d\omega_I}  dc \\
    &= \mathcal{N}(\omega_I \mid \mu_{I0}, \sigma_{I0}^2)\\
    &\times\int \frac{\mathcal{N}(y_I \mid \omega_I \cdot c, \sigma_I^2) \mathcal{N}(c \mid \mu_{T1}, \sigma_{T1}^2)}{\mathcal{N}(y_I \mid c \cdot \mu_{I0}, c^2\cdot\sigma_{I0}^2+\sigma_{I}^2)} dc
\end{align*}
\begin{figure*}
    \begin{center}
    \includegraphics[width=\textwidth]{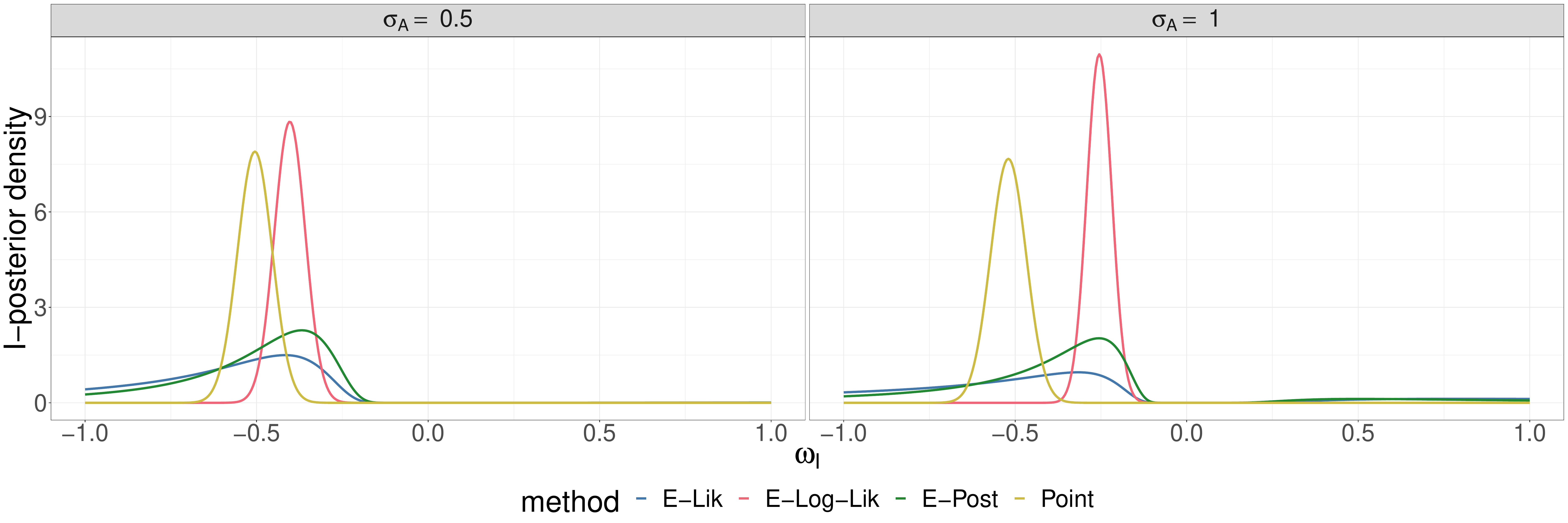}
    \caption{I-posterior distributions for the slope-only surrogate with normal priors/likelihoods. We used exemplary hyperparameters and data. We use the four UPs to compuete the I-posterior while varying $\sigma_A$.}
    \label{fig:slope_only_istep_posteriors}
    \end{center}
\end{figure*}
In \figref{fig:slope_only_istep_posteriors} we compare the four uncertainty propagation methods for exemplary parameters. To control for T-epistemic uncertainty we vary $\sigma_A$. 
% In general, we notice similar I-posteriors to the slope-intercept model (see \ref{subsection:linear_model}). 
In general, we notice similar I-posteriors to the slope-intercept model (see Main Section 3.1). 
Additionally, for large $\sigma_A$ we observe that both E-Lik and E-Post produce bimodal I-posteriors.

\subsubsection{Derivation E-Log-Lik}
We derive the Expected-Log-Likelihood result for the slope-only model:
\begin{align*}
    &p(\we \mid \ye, u=\text{E-Log-Lik})\\
    \propto{}& p(\we)\exp\left\{\int \log(p(\ye \mid \we, c))p(c \mid \cD_T) dc\right\} \\
    \begin{split}
        \propto{}& \cN(\omega_I \mid  \mu_{I1}, \sigma_{I1}^2) \\
        &\text{, with:}\\
        \sigma_{I1}^2 &= (\sigma_{I0}^{-2} +\se^{-2} (\mu_{T1}^2+\sigma_{T1}^2))^{-1}, \\
        \mu_{I1} &= \sigma_{I1}^2(\sigma_{I0}^{-2}\mu_{I0} + \se^{-2}\mu_{T1}\ye)
    \end{split}
\end{align*}
where we used:
\begin{align*}
    &\int \log(p(\ye \mid \we, c))p(c \mid \cD_T) dc\\
    ={}& \int \log(\mathcal{N}(y_I \mid \omega_I c, \sigma_{I}^2))\mathcal{N}(c \mid \mu_{T1}, \sigma_{T1})dc \\
    ={}& \int -\frac{1}{2\sigma_{I}^2}(\omega_I c-y_I)^2\mathcal{N}(c \mid \mu_{T1}, \sigma_{T1})dc \\
    ={}& -\frac{1}{2\sigma_{I}^2} \int (y_I^2 - 2y_I \omega_I c + c^2\omega_I^2)\\
    &\times\mathcal{N}(c \mid \mu_{T1}, \sigma_{T1})dc \\
    ={}& -\frac{1}{2\sigma_{I}^2} (y_I^2 - 2y_I \omega_I \nE[c]+\omega_I\nE[c^2])\\
    ={}& -\frac{1}{2\sigma_{I}^2} (y_I^2 - 2y_I \omega_I \mu_{T1}+\mu_{T1}^2 + \sigma_{T1}^2) \\
    \propto{}& -\frac{1}{2\sigma_{I}^2} (\mu_{T1}^2 + \sigma_{T1}^2)(\omega_I - \frac{\mu_{T1}y_I}{\mu_{T1}^2+\sigma_{T1}^2})^2 \\
\end{align*}
%\newpage
\subsection{Case Study 2: Logistic Model}
\subsection{T-posterior distribution}
\begin{figure*}
    \begin{center}
    \includegraphics[width=\textwidth]{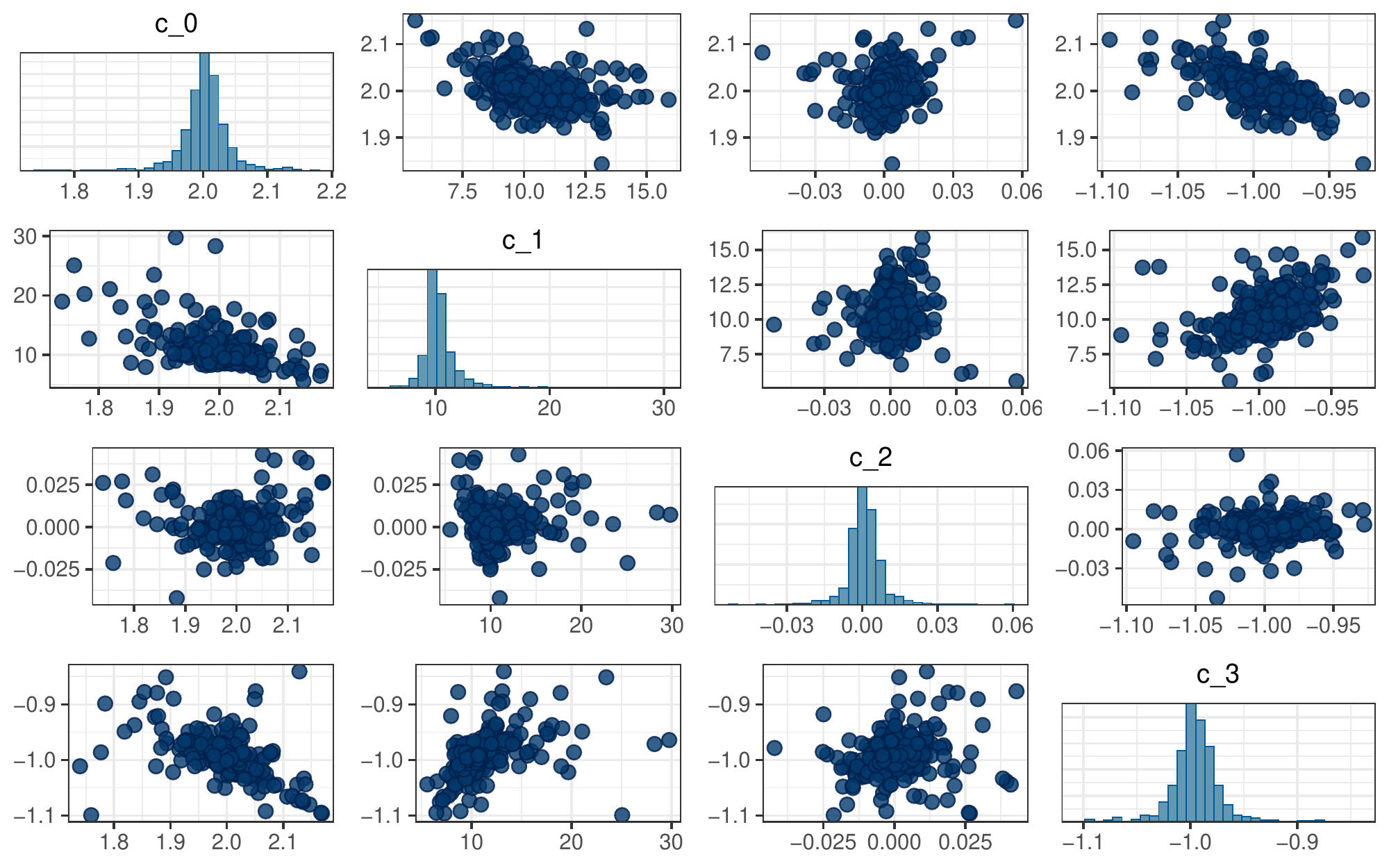}
    \caption{Pairs plot of the T-posterior draws of the logistic surrogate in Case study 2 for $N_T=7$.}
    \label{fig:pairs_plot_t_posterior}
    \end{center}
\end{figure*}

In \figref{fig:pairs_plot_t_posterior} show the T-posterior pairs plots using the logistic surrogate model ((same setup as in Main \secref{subsubsection:logistic_true}) for $N_T = 7$.
\subsubsection{PCE posterior distributions}\label{subsection:add_pce_post_densities}
\begin{figure*}
    \begin{center}
    \includegraphics[width=0.8\textwidth]{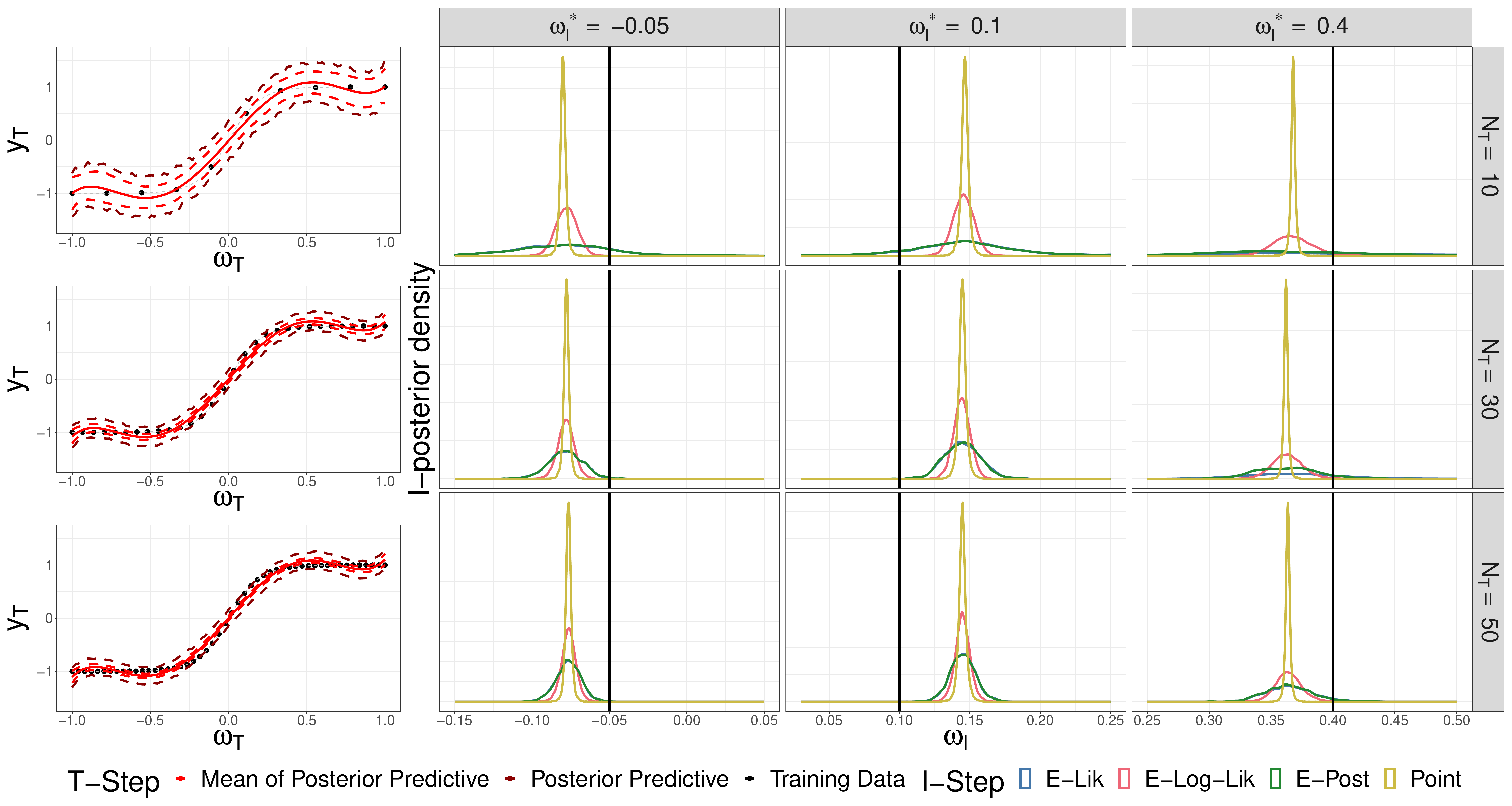}
    \caption{Additional results for T-epistemic UP ($\theta = c$) with PCE surrogate in Case study 2. Left: For $\Ns = \{10, 30, 50\}$ the training data set $\cD_T$ (black dots), the T-posterior predictive distribution and the mean of the T-posterior predictive distribution (dark red and red lines) is shown. Right: For the true inputs $\omega_I^* = \{-0.05, 0.1, 0.4\}$ (black vertical lines) we depict the I-posterior distributions for each Point, E-Lik, E-Post, and E-Log-Lik (colored lines).}
    \label{fig:i_steps_sigma_a_false_pce}
    \end{center}
\end{figure*}

\begin{figure*}
    \begin{center}
    \includegraphics[width=0.8\textwidth]{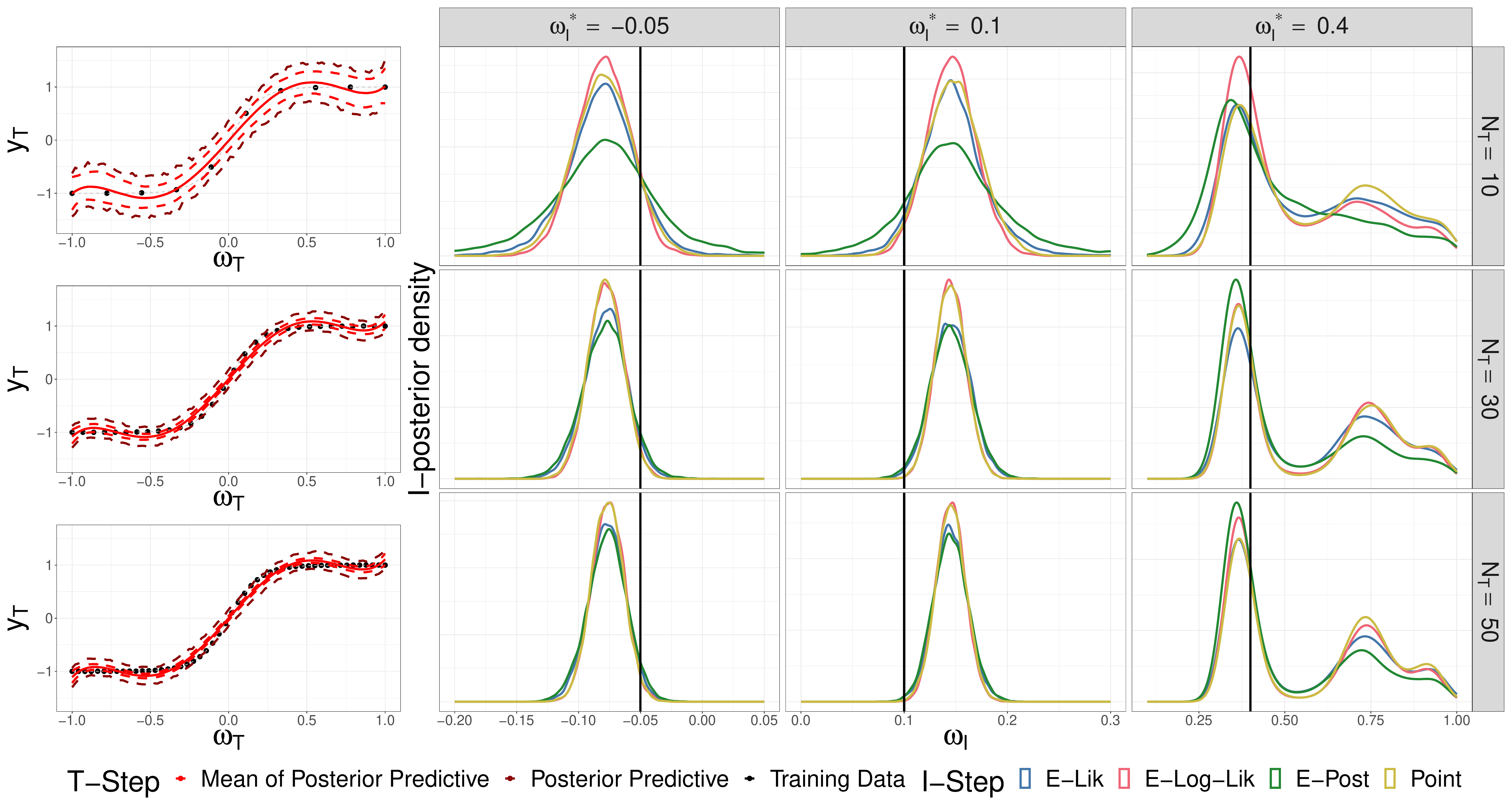}
    \caption{Additional results for T-epistemic and T-aleatoric UP ($\theta = (c, \sigma_A)$) with PCE surrogate in Case study 2. Left: For $\Ns = \{10, 30, 50\}$ the training data set $\cD_T$ (black dots), the T-posterior predictive distribution and the mean of the T-posterior predictive distribution (dark red and red lines) is shown. Right: For the true inputs $\omega_I^* = \{-0.05, 0.1, 0.4\}$ (black vertical lines) we depict the I-posterior distributions for each Point, E-Lik, E-Post, and E-Log-Lik (colored lines).}
    \label{fig:i_steps_sigma_a_true_pce}
    \end{center}
\end{figure*}

%We show I-posterior density plots using the PCE surrogate to approximate the logistic function (same setup as in \secref{subsubsection:logistic_pce}) for additional true input parameters $\omega_I^* = \{-0.05, 0.1, 0.4\}$. 
We show I-posterior density plots using the PCE surrogate to approximate the logistic function (same setup as in Main Section 3.2) for additional true input parameters $\omega_I^* = \{-0.05, 0.1, 0.4\}$. 
In \figref{fig:i_steps_sigma_a_false_pce} we propagate only epistemic uncertainty and in \figref{fig:i_steps_sigma_a_true_pce} we propagate both epistemic and aleatoric uncertainty.
\subsubsection{Computational Complexity}
\begin{figure*}
    \begin{center}
    \includegraphics[width=\textwidth]{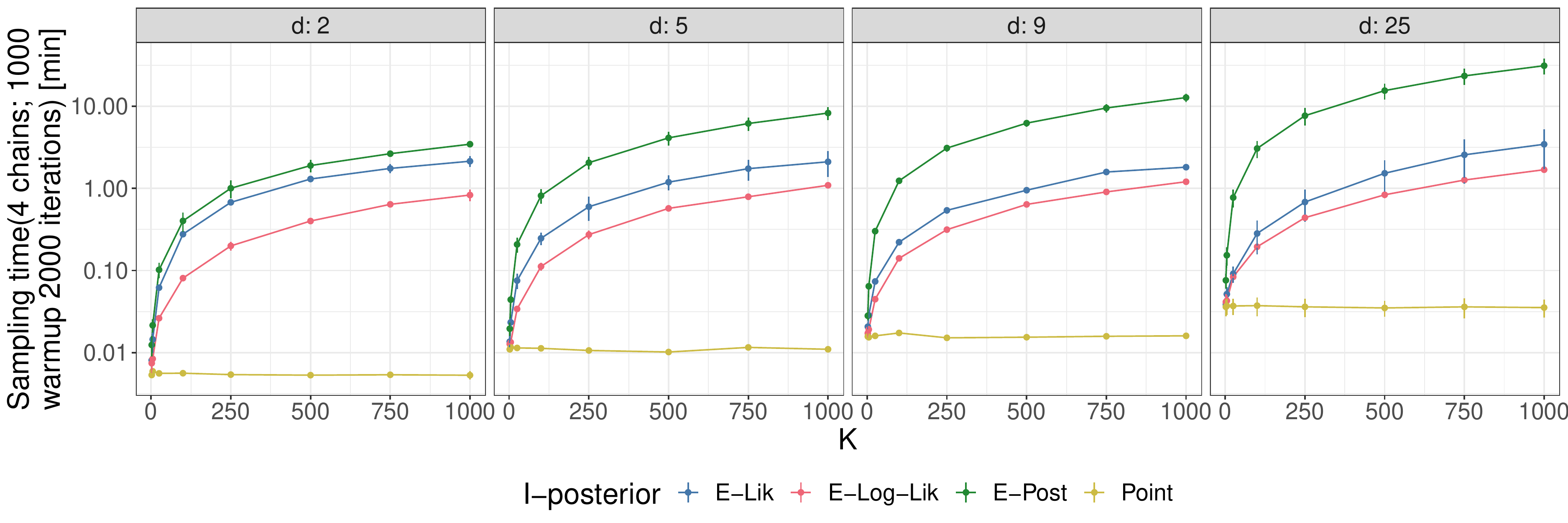}
    \caption{Sampling times estimated for Point, E-Lik, E-Log-Lik and E-Post shown on a logarithmic scale. For each method, we sample 4 chains with 1000 warmup iterations and 2000 total iterations. We vary the maximum degree of polynomials $d \in \{2, 5, 9, 25\}$ and the number of clusters $K \in \{2, 5, 25, 100, 250, 500, 750, 1000\}$.}
    \label{fig:sampling_log_times}
    \end{center}
\end{figure*}
% To evaluate the computational complexity of the different UP methods, we estimate their sampling times using a PCE surrogate approximating a logistic function (see \secref{subsubsection:logistic_pce}). 
To evaluate the computational complexity of the different UP methods, we estimate their sampling times using a PCE surrogate approximating a logistic function (see Main Section 3.2). 
For each UP method, we sample 4 chains with 1000 warmup iterations and 2000 total iterations. However, for a fair comparison, for E-Post we run for each model a full warmup phase with 1000 iterations, but sample only $2000/K$-times. The computations were not performed in parallel. We vary the maximum degree of polynomials $d \in \{2, 5, 9, 25\}$ and the number of clusters $K \in \{2, 5, 25, 100, 250, 500, 750, 1000\}$. We plot the times on a logarithmic scale in \figref{fig:sampling_log_times}.

\subsubsection{Parallelization}

To examine the effectiveness of parallelization, we measure the runtime for different numbers of workers, comparing the warmup and sampling times across the Point, E-Log-Lik, and E-Post methods. We use a PCE surrogate to approximate the logistic simulator, as described in Section 3.2.2. The Point method serves as a baseline, does not propagate uncertainty, and is not parallelized. To compute the I-posteriors using the E-Log-Lik and the E-Post method, we propagate $S=500$ T-posterior samples. 

We repeat the experiment three times and report the mean sampling time along with the standard deviation. The number of workers is varied across $\{1, 4, 16\}$. For each uncertainty propagation method, we run a single MCMC chain with 1000 warm-up and 1000 post-warmup iterations. Fig. \ref{fig:parallelization} presents the results on a logarithmic scale.

This plot confirms that the runtime for both E-Log-Lik and E-Post decreases as the number of workers increases. Additionally, we observe that E-Post benefits more from parallelization than E-Log-Lik, which aligns with the expected degree of parallelizability of these methods as discussed in Section 2.2.6. Due to implementation constraints, we exclude E-Lik from this analysis, as its use of the \texttt{log-sum-exp} trick for numerical stability currently limits straightforward parallelization. However, given the similarities in implementation between E-Lik and E-Log-Lik, we expect that the conclusions drawn for E-Log-Lik also apply to E-Lik.

\begin{figure}
    \begin{center}
    \includegraphics[width=0.45\textwidth]{figures/runtime_number_workers_pce.pdf}
    \caption{Effect of parallelization on the warmup and sampling times. The runtimes for E-Log-Lik, E-Post, and Point are shown on a logarithmic scale.}
    \label{fig:parallelization}
    \end{center}
\end{figure}

\end{document}